\documentclass[lettersize,journal,twoside,article]{IEEEtran}

\usepackage{amsmath,amsfonts}
\usepackage{algorithmic}
\usepackage{algorithm}
\usepackage{array}
\usepackage[caption=false,font=normalsize,labelfont=sf,textfont=sf]{subfig}
\usepackage{textcomp}
\usepackage{stfloats}
\usepackage{url}
\usepackage{verbatim}
\usepackage{graphicx}
\usepackage{cite}

\usepackage{multirow}
\usepackage{hyperref}
\usepackage{pmboxdraw}
\usepackage[table,xcdraw]{xcolor}
\usepackage{nicematrix}
\usepackage[flushleft]{threeparttable}
\usepackage{enumitem}
\usepackage{orcidlink}
\usepackage{soul}
\usepackage{makecell}

\newcommand{\equref}[1]{Eq.~\(\ref{#1}\)}
\newcommand{\figref}[1]{Figure~\ref{#1}}
\newcommand{\secref}[1]{Section~\ref{#1}}
\newcommand{\tabref}[1]{Table~\ref{#1}}

\usepackage[export]{adjustbox}

\begin{document}


\title{Universal Embedding Function for Traffic Classification via QUIC Domain Recognition Pretraining: A Transfer Learning Success}

\author{Jan Luxemburk\orcidlink{0000-0003-0879-0054},
        Karel Hynek\orcidlink{0000-0002-8281-618X},
        Richard Plný\orcidlink{0000-0002-0544-8424},
        and~Tomáš Čejka\orcidlink{0000-0001-7794-9511}
\thanks{Accepted for publication in \textit{IEEE Transactions on Network and Service Management}, DOI: 10.1109/TNSM.2025.3642984}
\thanks{This research was funded by the Ministry of Interior of the Czech Republic, grant No. VJ02010024: \textit{``Flow-Based Encrypted Traffic Analysis''} and also supported by the Grant Agency of the CTU in Prague, grant No. SGS23/207/OHK3/3T/18. Computational resources were provided by the e-INFRA CZ project (ID:90254), supported by the Ministry of Education, Youth and Sports of the Czech Republic.}
\thanks{Jan Luxemburk, Richard Plný, and Karel Hynek are with the Faculty of Information Technology, Czech Technical University in Prague, Prague 160 00, Czech Republic, and also with the CESNET association, Prague 160 00, Czech Republic. \textit{Corresponding author: Jan Luxemburk (luxemburk@cesnet.cz)}}
\thanks{Tomáš Čejka is with CESNET association, Prague 160 00, Czech Republic.}
}

% \markboth{IEEE TRANSACTIONS ON NETWORK AND SERVICE MANAGEMENT, VOL. XX, NO. X, XXXX XXXX}{Luxemburk \MakeLowercase{et al.}: Universal Embedding Function for Traffic Classification}

\maketitle

\bstctlcite{IEEEexample:BSTcontrol}

\begin{abstract}
Encrypted traffic classification (TC) methods must adapt to new protocols and extensions as well as to advancements in other machine learning fields. In this paper, we adopt a transfer learning setup best known from computer vision. We first pretrain an embedding model on a complex task with a large number of classes and then transfer it to seven established TC datasets. The pretraining task is recognition of SNI domains in encrypted QUIC traffic, which in itself is a challenge for network monitoring due to the growing adoption of TLS Encrypted Client Hello. Our training pipeline---featuring a disjoint class setup, ArcFace loss function, and a modern deep learning architecture---aims to produce universal embeddings applicable across tasks. A transfer method based on model fine-tuning surpassed SOTA performance on nine of ten downstream TC tasks, with an average improvement of 6.4\%. Furthermore, a comparison with a baseline method using raw packet sequences revealed unexpected findings with potential implications for the broader TC field. We released the model architecture, trained weights, and codebase for transfer learning experiments.

\end{abstract}


\section{Introduction}

In this paper, we propose a universal embedding (mapping) function that transforms packet sequences into an embedding vector space. The core idea is to map similar packet sequences close to each other in the embedding space while keeping dissimilar ones far apart. The embedding function serves as a feature extractor, enabling a nearest neighbors (k-NN) classifier to make predictions. As our focus is on encrypted traffic, we utilize the standard input representation unaffected by encryption: packet size, direction, and inter-packet time of the first \textit{N} packets. We deliberately avoid using payload as a model input due to its limited value in analyzing initial handshakes of encrypted protocols, a task we consider better suited to protocol dissectors and pattern-matching. Additionally, in deployment scenarios where inference is not performed directly on monitoring probes, transmitting payload data introduces both privacy and performance issues.

Building on our previous research in fine-grained traffic classification (TC) for TLS~\cite{Luxemburk2023TLS} and QUIC~\cite{Luxemburk2023QUIC}, we design and train the embedding function using the CESNET-QUIC22~\cite{Luxemburk2023QUICDataset} dataset. This dataset includes Server Name Indication (SNI) domains as labels, enabling a classification task focused on inferring exact domain names from packet sequences. This domain recognition task serves two important roles in this paper. First, the growing adoption of the Encrypted Client Hello (ECH) extension has made domain recognition increasingly important. In combination with TLS 1.3, ECH removes all plaintext handshake fields that have traditionally enabled visibility into encrypted traffic. Our solution, which relies solely on packet sequences and is unaffected by ECH, is able to infer the correct domain name in 94.83\% of cases, even when evaluated on test domains entirely disjoint from the embedding function's training set. Second, domain recognition is well suited as a pretraining task due to its complexity, large number of classes, and straightforward labeling process. Transfer learning leverages models trained on one task to adapt them for a different but related task, under the assumption that certain knowledge~(i.e., extracted features, learned traffic patterns and characteristics) is shared and transferable. This approach is widely used in computer vision and natural language processing, where the typical experimental pipeline involves fine-tuning large models pretrained on public datasets.

To evaluate the transfer learning approach, we tested our pretrained embedding function on seven datasets: \mbox{ISCXVPN2016}~\cite{dataset_ISCXVPN2016}, MIRAGE19~\cite{dataset_MIRAGE19}, MIRAGE22~\cite{dataset_MIRAGE22}, UTMOBILENET21~\cite{dataset_UTMOBILENET21}, UCDAVIS19~\cite{dataset_UCDAVIS19}, CESNET-TLS22~\cite{Luxemburk2023TLS}, and AppClassNet~\cite{dataset_AppClassNet}, presenting ten downstream TC tasks in total. We evaluated three transfer techniques: \textit{(a)}~using fixed pretrained embeddings with a k-NN classifier, \textit{(b)}~using fixed pretrained embeddings with a linear classifier (also known as linear probing), and \textit{(c)}~fine-tuning the embedding function on each downstream task. To isolate the contribution of the transfer learning, we also compared against training the same model from scratch on the downstream tasks.

% \IEEEpubidadjcol

The results are promising: the fine-tuning approach proved to be the best, surpassing state-of-the-art (SOTA) performance on nine downstream tasks and outperforming training from scratch on eight. A k-NN classifier in the learned embedding space also exceeded SOTA performance, although the gains were more modest compared to fine-tuning. This offers a viable alternative when fine-tuning or training from scratch is not feasible, for example due to insufficient labeled data for the downstream task. Finally, our experiments revealed an intriguing finding: a k-NN classifier using L1 distance on the first 10 packet features---referred to as the input-space baseline---also performed quite well across all datasets.

\medskip
\noindent
\textit{Main contributions}

\textit{(i)} We designed and developed \textbf{an embedding function for packet sequences to serve as a foundation for classifiers of encrypted network traffic.} It is based on a neural network architecture \texttt{30pktTCNET\_256} that combines existing, well-proven techniques, such as ResNet-like convolutional blocks, packet feature embeddings, and Generalized Mean Pooling (GeM)~\cite{Radenovic2018GeM}. For initializing packet feature embeddings, we adopted Piecewise Linear Encoding (PLE)~\cite{gorishniy2022embeddings}, a method that outperforms random initialization and has not previously been used in the TC domain. Although the architecture is largely based on established techniques---with the exception of PLE---we rigorously tuned all its components to uncover the true limits of CNN-based processing of packet sequences.

\smallskip
\textit{(ii)} We present \textbf{a domain recognition solution that enables domain-level visibility into encrypted traffic, even in the presence of TLS 1.3 and ECH.} Using a k-NN classifier in the embedding space, our method achieves 94.83\% accuracy and 79.35\% recall under a class-disjoint evaluation setup. We further show that domain recognition serves as an effective pretraining task, as the learned embeddings generalize across ten downstream TC tasks.

\smallskip
\textit{(iii)} With the fine-tuning transfer method, we \textbf{surpass the SOTA performance on nine of the ten downstream TC tasks, achieving a remarkable average improvement of~6.4\%}. Compared to training from scratch, fine-tuning pretrained weights provides a 2.1\% average gain, empirically confirming the benefits of transfer learning.

\medskip
\noindent
The paper is organized as follows: \secref{sec:related-work} provides a review of related research. \secref{sec:experimental-setup} describes the experimental setup used to develop the embedding function, covering the dataset, data preparation, training loop, loss function, and deep learning (DL) architecture. \secref{sec:domain-recognition-results} presents the results of the domain recognition task, along with ablation studies examining parts of the solution. \secref{sec:transfer-learning} describes the transfer of the trained embedding function to seven additional TC datasets and discusses the achieved results. \secref{sec:limitations} addresses limitations, and \secref{sec:conclusion} concludes the paper by summarizing key contributions and outlining future directions.

\section{Related work}
\label{sec:related-work}
We begin by outlining how transfer learning has been applied within the TC domain. Next, we review influential papers on representation learning for TC, highlighting methodological similarities to our approach and key differences. We deliberately omit detailed discussion of studies that rely heavily on payload data, focusing instead on methods that operate on packet sequences or on representations derived from them.

\subsection{Transfer learning in traffic classification }
The objective of transfer learning is to capitalize on the knowledge encoded in a model $M_{0}$, learned while solving a task $T_{0}$, and adapt it to a different target task $T_{target}$. The key intuition is that the closer the relationship between $T_{0}$ and $T_{target}$, the more effectively the learned representations (i.e., features) can be reused and transferred. Depending on the application, the label spaces of $T_{0}$ and $T_{target}$ may be disjoint (e.g., when pretraining on a pretext task), partially overlapping (e.g., when adapting to new classes), or identical (e.g., when adapting to a new operational environment). Transfer learning has gained significant traction in network classification and is applied in various scenarios discussed below.

\textit{1)~Cross-dataset: pretraining on pretext tasks before fine-tuning for downstream objectives.} ET-BERT~\cite{lin2022Etbert} and YaTC~\cite{Zhao2023Yatc} exemplify this strategy, leveraging large unlabeled datasets to pretrain transformer models in an NLP-like fashion using tokenization and masking. However, both operate on packet payloads and are thus outside the scope of detailed comparison in this work. Rezaei et al.~\cite{Rezaei2020Multitask} pretrained a 1D-CNN on QUIC packet sequences via a pretext task targeting flow bandwidth and duration prediction, and subsequently transferred it to web service classification. This pretraining improved performance relative to direct training on web services; nevertheless, jointly training all three tasks---bandwidth, duration, and web service---yielded the strongest performance. 

\textit{2)~New-class adaptation: transferring knowledge from frequent to rare classes in incremental learning and few-shot learning (FSL).} When the main constraint is a limited number of samples, transfer learning can serve as an alternative to, or a component within, various FSL approaches. Bovenzi et al.~\cite{Bovenzi2024Attack} studied IoT attack detection across four datasets and found that FSL methods outperform transfer learning. However, subsequent works by Monda et al.~\cite{Monda2024Few, Monda2024Botnet}, focusing on intrusion detection and fast adaptation to new attacks, reported the opposite result: transfer learning methods were superior. In particular, the \textit{Rethinking Few-Shot} method of Tian et al.~\cite{Tian2020RFS} achieved excellent performance using a base model trained in a standard supervised fashion, with additional self-distillation to enhance the learned representations. The \textit{Rethinking Few-Shot} method highlights that a strong base model can offer performance comparable to, or better than, standard FSL meta-learning approaches while enabling a simpler and arguably more straightforward adaptation to new classes. Tong et al.~\cite{Tong2025Adaptation} proposed a feature-alignment approach for encrypted traffic classification. In their problem setup, the source domain provides abundant labeled data, whereas the target domain has many samples but few or no labels and includes additional classes. Their method introduces an auxiliary optimization objective, termed the smooth characteristic function, which encourages the feature distributions of the two domains to align during training. This alignment facilitates classification of previously unseen classes in the target domain, although the approach requires target-domain data to be available during base-model training. 

\textit{3)~Cross-network: adapting models trained in one environment to operate in another.} Unlike most approaches discussed so far, which aim to learn embeddings of individual network connections, Gioacchini et al.~\cite{Gioacchini2024Cross} learn embeddings of hosts. They employ DarkVec~\cite{Gioacchini2021Darknet}, a Word2Vec-based method that learns host embeddings from co-occurrence patterns---for example, when hosts contact the same server ports at the same time. The authors then study two transfer settings: \textit{(a)} adapting host embeddings from a provider network to different, potentially label-scarce client networks, and \textit{(b)} transferring knowledge learned from honeypots to scanners observed by a network telescope.

\subsection{Contrastive learning \& Augmentations}
Guarino et al.~\cite{Guarino2023Many} focused on the task of finding better representations for TC that generalize across tasks, which is in line with the goals we set up for this work. The authors compared transfer-, meta-, and contrastive-learning approaches on MIRAGE19 and AppClassNet datasets. Both datasets were partitioned into training, validation, and test sets with disjoint classes, similar to how we partition domain names for the domain recognition task. However, their partitioning approach was based on class frequencies: the training set contained the most frequent classes, while the test set included the least frequent ones. In contrast, we used random partitioning independent of class frequencies. One of the findings of Guarino et al. is that supervised contrastive learning produces the best representations overall. Our work supports this finding, as there are notable similarities between the loss function used in our work, ArcFace, and the contrastive loss function used by Guarino et al:~SupCon~\cite{khosla2020supervised} extended with class centers and cosine distance. Both loss functions operate by pulling similar samples toward their class centers using cosine similarity. The key distinction is that ArcFace introduces angular margins, which enhance class separation in the embedding space.

Wang et al.~\cite{wang2024augmentation} presented a benchmark of data augmentations, evaluating 18 of them on three datasets: MIRAGE19, MIRAGE22, and a private one. They utilized the standard packet sequence representation, consisting of the sizes, times, and directions of the first 20 packets. The training pipeline featured two stages: a contrastive self-supervised phase, where augmented versions of the same sample were pulled together, followed by a supervised phase, where a classification head was trained on the extracted features. Wang et al. also experimented with a class-weighted sampler to achieve perfect class balancing in each training epoch but found no success with it. In contrast, our semi-balancing technique, detailed in~\secref{sec:training-sampler}, proved to be beneficial in our experiments. The differing outcomes may be due to the use of perfect class balancing, whereas we employed the $\lambda_{sampler}$ parameter to control the strength of the balancing effect. We used the best classification accuracies reported by Wang et al. on the MIRAGE19 and MIRAGE22 datasets for SOTA comparison in~\secref{sec:transfer-learning}. Xie et al.~\cite{Xie2023_Rosetta} carefully designed three TCP-aware data augmentations that mimic real network dynamics, such as varying packet loss rates, retransmission timeouts, or the interplay between RTT and MTU in packet buffering. Although their objective---to train encrypted traffic classifiers that are more robust under changing network conditions---aligns with our focus on general traffic representations, we chose not to incorporate data augmentations into our pipeline due to the increasing complexity of our proposed approach.

Finamore et al.~\cite{finamore2023replication} conducted a comprehensive evaluation of data augmentations of the FlowPic~\cite{Shapira2021_FlowPic} representation, which is a 2D histogram that captures the evolution of packet sizes over time. The authors replicated an earlier study~\cite{Horowicz2022flowpic}, reproducing most of the original results while also incorporating three additional datasets. The most relevant contribution for our work is the release of the \textit{tcbench} open-source framework, which we use for transfer learning evaluation. Additionally, the best classification accuracies reported on the UCDAVIS19 and UTMOBILENET21 datasets are used for SOTA comparison.

\section{Experimental setup}
\label{sec:experimental-setup}
\begin{figure*}[t]
    \centering
    \includegraphics[width=0.76\textwidth]{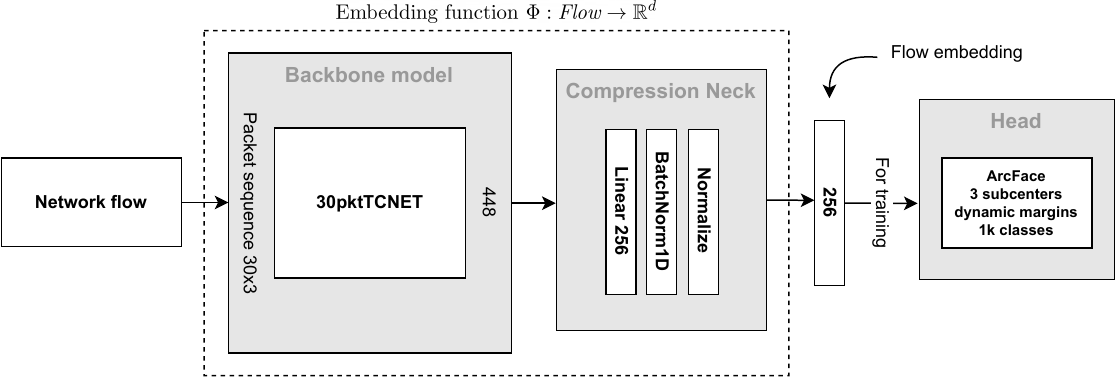}
    \caption{A complete processing pipeline starting with network flows as input. The embedding function $\Phi$, which is implemented as a neural network, maps flows into a 256-dimensional vector space. The visualized ArcFace head is used during training to optimize the neural network, which is composed of a backbone model and a compression neck.}
    \label{fig:embedding-function}
\end{figure*}

The overall domain recognition experimental setup is designed as a retrieval task for finding the most similar network flows, analogous to image retrieval tasks in computer vision. The SNI-based classes are divided into three disjoint sets: training, validation, and test. The training domain set is used to train a neural network, which serves as the embedding function that learns vector representations of network flows. This embedding function, denoted as $\Phi$, is formalized in~\equref{equ:embedding-function} and illustrated in~\figref{fig:embedding-function}.
\begin{equation}
\label{equ:embedding-function}
\Phi: \textit{Flow} \to \mathbb{R}^d,\,d = \textit{embedding size}
\end{equation}
The validation domain set is used to measure performance during training, select the best model, and for finding the best configuration of hyperparameters. The test domain set is reserved for measuring and reporting the final metrics. The training, validation, and test domain sets are \textbf{disjoint}, so this setup pushes for strong generalization capabilities of the embedding function. It must learn patterns and extract traffic characteristics that remain useful for domains not seen during training. We based our experiments on the CESNET-QUIC22~\cite{Luxemburk2023QUICDataset} dataset, which is described in the next section.

\subsection{CESNET-QUIC22 dataset}
\label{sec:dataset}
The CESNET-QUIC22~\cite{Luxemburk2023QUICDataset} dataset includes four weeks of traffic captured at the monitoring points of the CESNET network, which is the national research and education network of the Czech Republic. The dataset consists of 153 million anonymized network flows enriched with various traffic features suitable for the classification of encrypted traffic. For this work, the relevant features are the packet sequences and SNI domains. The SNI domain is extracted from the Server Name Indication extension transmitted during the QUIC handshake. The packet sequences include packet sizes, inter-packet times, and packet direction of the first 30 packets. Prior work has shown that this packet sequence length is generally sufficient for accurate classification~\cite{Luxemburk2023QUIC, Luxemburk2023TLS}. For further details on the dataset and its collection process, including the software used, sampling methods, flow export timeouts, and other relevant aspects, please refer to the original data article~\cite{Luxemburk2023QUICDataset}.

The dataset includes 102 service classes, where each \textit{service} groups one or more domain names under a single label. However, since our objective is to predict individual domains, we do not use these predefined service labels. Instead, we utilized the exact SNI domains that are available in the dataset. Moreover, since the dataset contains general background traffic, the domains are not limited to the 102 service classes but represent all domains observed within the CESNET network.

In our previous experiments~\cite{Luxemburk2023QUIC} with CESNET-QUIC22, we observed significant data drift in the traffic from the first two weeks of the dataset. In this work, we want to focus on evaluating the embedding function for domain recognition and transfer learning, without introducing extra complexities of the data drift. Therefore, we based our experiments on the third week of the dataset {(W-2022-46)} and used 33.7 million samples from this week for training, hyperparameter tuning, and the final evaluation of the proposed solution.

\subsection{Experimental pipeline}
\begin{figure}[t]
    \centering
    \includegraphics[width=0.84\columnwidth]{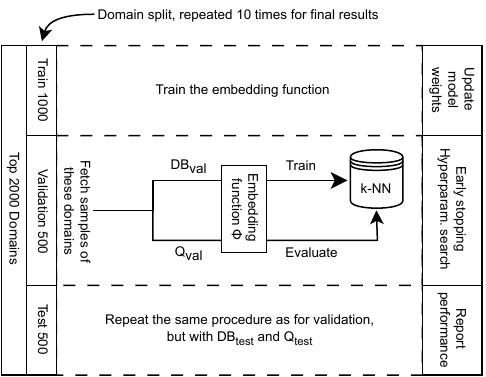}
    \caption{An overview of the experimental setup, highlighting the purpose of the disjoint domain split along with the database and query preparation for validation and testing.}
    \label{fig:experimental-setup-overview}
\end{figure}

The following subsections describe the individual steps of the experimental setup, starting with domain preprocessing, followed by data preparation, training and validation, and finally model selection. An overview of the full pipeline is shown in~\figref{fig:experimental-setup-overview}.

\subsubsection{Domain preprocessing and train/val/test split}
\label{sec:domain-preprocess}
The first step was to preprocess SNI domain names into class labels. We keep subdomains up to the fourth level and strip the rest (\texttt{a.tile.openstreetmap.org} is a fourth-level domain). Some domains contain a random string or parts related to locations or numbering. For example, the aforementioned openstreetmap domain has two "sister" domains \texttt{[b,c].tile.openstreetmap.org}. We decided to group such domains into a single class with the help of regexes. Another example would be \texttt{europe-west1-gcp.api.snapchat.com} and \texttt{us-east4-gcp.api.snapchat.com}, both remapped to a single class \texttt{\$LOC-gcp.api.snapchat.com}. In total, we created 40 regexes that remap thousands of domains with random parts into corresponding unified \textit{domain classes}.

After this domain preprocessing, we selected the 2000 most frequent \textit{domain classes} and divided them randomly into three subsets: 1000 for training, 500 for validation, and 500 for testing. These 2000 domains account for 99.38\% of the total flows in the dataset's third week.

\subsubsection{Database and query preparation}
\label{sec:database-preparation}
Next, we prepare databases and query samples for validation and test domain sets. A database serves as a mini-training set for a k-NN classifier, which is then tested on query samples to measure performance. The exact same process is used for validation and test domains, and we will describe it for validation.

Validation samples (those having one of the 500 validation domains) are split into database and query parts. This split is random and stratified, meaning the class frequencies are preserved in both parts. We set the query part to contain one million samples and leave the rest for building the database. Out of these remaining samples, we randomly select one million to be included in the database. However, this second database sampling is not uniform but instead we soften the class imbalances. We set the weight of a sample belonging to class $C$ to $N_{C}^{-\lambda_{db}}$, where $N_C$ is the $C$ class frequency and $\lambda_{db} > 0$ a parameter controlling the strength of the balancing effect. We ended up using $\lambda_{db} = \frac{1}{2}$, meaning the weight formula is $\frac{1}{\sqrt{N_C}}$. This method for the selection of database samples prioritizes classes that are less frequent at the expense of the most frequent ones. To summarize, for both validation and test domain sets, we selected one million query samples ($Q_{val}$, $Q_{test}$) that follow the dataset's original class distribution, and we created a database ($DB_{val}$, $DB_{test}$) from one million samples with a more balanced class distribution. 

\subsubsection{Training, validation, and ranking}
\label{sec:ranking}
After preprocessing domains, splitting them into disjoint sets, and preparing databases and query samples, we can start training the $\Phi$ embedding function using the training domain set. The training loop runs for 30 epochs, measuring validation metrics each two epochs. The training loop is described in detail in~\secref{sec:training-loop}. The validation is performed as a similarity search in the embedding space. For each $Q_{val}$ sample, we find the most similar $DB_{val}$ samples, a process we refer to as database ranking. We limit the ranking to the closest 20 samples and compute several metrics that are described in~\secref{sec:metrics} to measure the quality of the embeddings. Each time, validation reuses the same $DB_{val}$ and $Q_{val}$ samples---what changes are the embeddings that are recomputed with model weights from the current epoch.

To compute distances in the embedding space, we use cosine similarity, which is a metric calculated as the dot product of L2-normalized vectors. When applied to L2-normalized vectors, cosine similarity produces the same ranking as the well-known Euclidean distance. To efficiently compute cosine similarities between all query samples and all database samples, we used the \textit{faiss} library~\cite{johnson2019billion} specifically designed for efficient similarity search and clustering of vectors.

\subsubsection{Model selection and final evaluation}
We chose the macro-average validation recall as our main objective, favoring models that perform well on all validation domains with equal importance (macro-averaging disregards sample count per domain). After the 30 training epochs, the model from the epoch with the best validation recall is saved and evaluated on the test domain set. This final evaluation is identical to the validation process but is using the test query samples $Q_{test}$ and test database $DB_{test}$.

\subsection{DL architecture}
\label{sec:dl-architecture}
\begin{figure*}[t]
    \centering
    \includegraphics[width=0.76\textwidth]{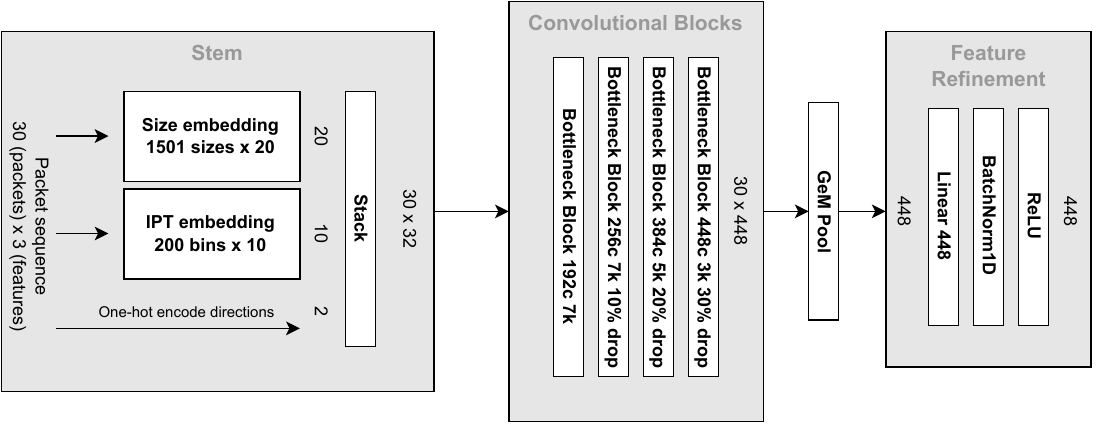}
    \caption{The architecture of the \texttt{30pktTCNET} backbone model consists of four main components: a stem, convolutional blocks, global pooling, and feature refinement. The main processing is done in the convolutional blocks, which include four Bottleneck Residual Blocks described in detail in~\figref{fig:bottleneck-block}. Each block has a different configuration of the following parameters: the number of output channels (e.g., 256c), kernel size (e.g., 7k), and dropout rate.}
    \label{fig:backbone-model}
\end{figure*}

We based the neural network architecture on our previous works on TLS and QUIC classification~\cite{Luxemburk2023TLS, Luxemburk2023QUIC}, but due to substantial modifications, we describe it here in detail. The architecture is single-modal and designed to process fixed-sized packet sequences ($N = 30$) with the following features: packet sizes, inter-packet times (IPT), and directions. Network flows with fewer than $N$ packets are padded with zeroes. The network is visualized in Figure~\ref{fig:backbone-model}. It follows the standard architecture of modern CNNs and consists of four main components: a stem, convolutional blocks, global pooling, and a feature refinement block.

\subsubsection{Stem}
\label{sec:stem}
The purpose of our network stem is to embed packet features into $R$-dimensional vectors to prepare them for subsequent processing with convolutions. This is achieved using two PyTorch \textit{Embedding}\footnote{\url{https://pytorch.org/docs/stable/generated/torch.nn.Embedding.html}.} layers---one for embedding of packet sizes and another for IPT. Each \textit{Embedding} layer contains a learnable matrix with the shape \textit{number of embeddings} $\times$ \textit{embedding size}, where each row represents the embedding vector for a specific value. Packet sizes range from 0 to 1500, resulting in 1501 embeddings. For IPT, we first bin the values into 200 bins and then use the index of a bin as the input for the \textit{Embedding} layer. Packet directions are one-hot encoded, which we found to be more effective than the traditional $\pm1$ encoding scheme. We set the embedding size to 20 for packet sizes and 10 for IPT. Thus, the stem outputs data in a shape (30 $\times$ 32)\footnote{The batch size is omitted from all data shapes discussed in~\secref{sec:dl-architecture}.}, where 30 corresponds to the packet sequence length, and 32 represents the combined embedding vector (20 for packet sizes, 10 for IPT, and 2 for directions).

The \textit{Embedding} layers for packet sizes and IPTs were initialized using the Piecewise Linear Encoding (PLE) method proposed in~\cite{gorishniy2022embeddings}, rather than the default random initialization. PLE creates initial embeddings structured as bins, where each bin corresponds to a segment of the feature's range (hence the name "piecewise"). Within each bin, linear relationships are preserved, maintaining the inherent ordering of numerical features. During training, the embeddings are optimized alongside the other model weights to adapt to the data. While embedding packet features before convolutional processing is not a novel concept---having been employed, for instance, by Nascita et al.~\cite{Nascita2021FirstEmbeddings, Nascita2023Embeddings}---the PLE initialization technique has not yet been used in the TC domain. 

\subsubsection{Convolutional blocks}
\begin{figure}[t]
    \centering
    \includegraphics[width=0.35\textwidth]{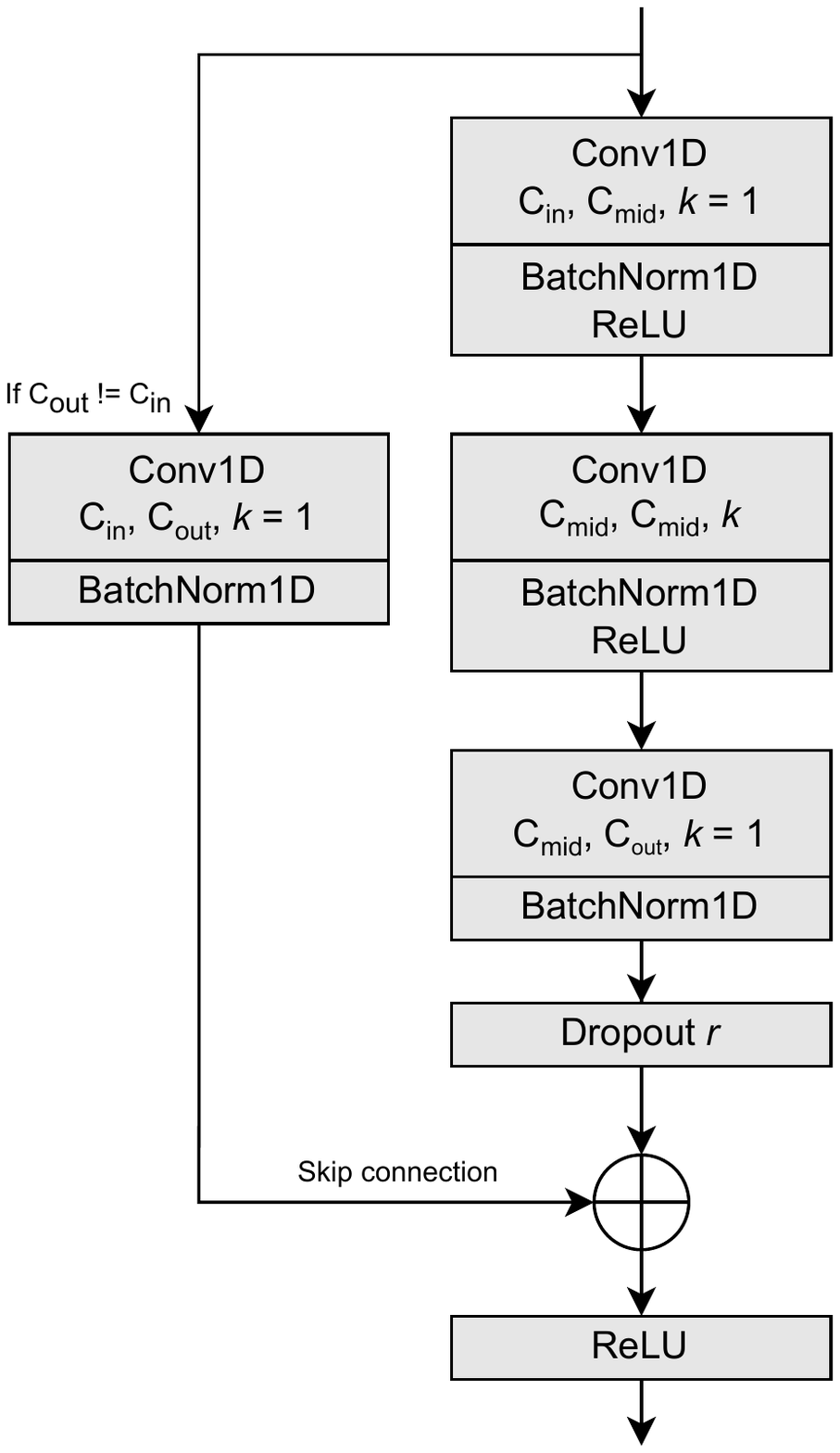}
    \caption{The diagram of Bottleneck Residual Block. $k$: the kernel size of the main convolution, $C_{out}$: the number of output channels, $r$: the dropout rate. The number of channels of the main convolution $C_{mid}$ is set to $\frac{C_{out}}{4}$. All convolutions use a stride of 1 and automatic padding to ensure that the spatial dimension is kept intact. Convolutions do not use biases.}
    \label{fig:bottleneck-block}
\end{figure}

The core processing in terms of feature extraction and parameter count is done with convolutions, implemented as four residual blocks adapted from the popular ResNet architecture. Specifically, we use Bottleneck Residual Block visualized in~\figref{fig:bottleneck-block}, which was proposed in~\cite{He2016DeepResidual}. This block design reduces the number of parameters while preserving representational power. The term "bottleneck" refers to the temporary reduction of channels with a 1$\times$1 convolution before the main convolution, followed by their restoration (or increase) afterward. Each block is defined by the following parameters: the main convolution kernel size $k$, the number of output channels $C_{out}$, the dropout rate, and the bottleneck ratio defining the reduction of channels for the main convolution (we use a common setting of $\frac{1}{4}$). The main convolution operation uses automatic padding to ensure that the spatial dimension (i.e., the length of packet sequences) remains unchanged. For the same purpose, we also use a stride of 1 in all convolutions, as we found that reducing the spatial dimension with a stride led to decreased performance. The four bottleneck blocks use different parameter settings: the number of output channels increases (192, 256, 384, 448), the kernel sizes decrease (7, 7, 5, 3), and the dropout rates progressively rise (0\%, 10\%, 20\%, 30\%). Overall, the convolutional blocks process packet embeddings of shape (30 $\times$ 32) and produce feature maps of shape (30 $\times$ 448).

\subsubsection{Global pooling}
The next component of the network is a global pooling operation, which aggregates each feature map along the spatial dimension (length 30) into a single scalar per channel, producing an output of size (1 $\times$ 448). Common pooling methods include either averaging or taking the maximum of the values. We found maximum pooling to perform better than average pooling; however, our final choice was Generalized Mean Pooling (GeM)~\cite{Radenovic2018GeM}. GeM includes a parameter $p$ that enables interpolation\footnote{The exact GeM formula is $\left( \frac{1}{|X_c|} \sum\limits_{x \in X_c} x^p \right)^{\frac{1}{p}}$, where $X_c$ denotes a single channel from the output of the convolutional blocks.} between maximum pooling ($p \to \infty$) and average pooling ($p = 1$). The parameter $p$ can be a fixed value or trained along with the other model weights. We initialized $p$ to 3 and optimized it during training.

\subsubsection{Feature refinement \& Compression neck}
The output of the GeM pooling is passed through a \textit{feature refinement block}---a simple sequence of Linear, BatchNorm, and ReLU layers. This \textit{feature refinement block} preserves the shape of the features, resulting in an output size of 448.

Up to this point, the defined neural network architecture can be used for standard classification tasks; adding one extra Linear classification layer with a shape (448 $\times$ \textit{number of classes}) would do the job. However, our goal is to produce embeddings of network flows. Thus, as the final part of the network, we add a \textit{compression neck} that is composed of a Linear layer with a shape (448 $\times$ 256), BatchNorm, and a vector L2-normalization operation. The \textit{compression neck} excludes a ReLU activation function on purpose, as its task is to compress features into the desired embedding size of 256 without introducing additional non-linear transformations. 

We want to distinguish the backbone part---the stem, convolutional blocks, and the \textit{feature refinement block}---from the entire neural network. We refer to this backbone as \texttt{30pktTCNET} (visualized in~\figref{fig:backbone-model}), while the complete model representing the $\Phi$ embedding function is denoted as \texttt{30pktTCNET\_256} (visualized in~\figref{fig:embedding-function}). With the final hyperparameter configuration, \texttt{30pktTCNET\_256} has one million trainable parameters.

\subsection{Training loop}
\label{sec:training-loop}
This section outlines the training loop that we used to optimize the neural network, describing the training sampler, loss function, optimizer, and learning rate (LR) scheduler.

\subsubsection{Training sampler}
\label{sec:training-sampler}
The training loop consists of 30 epochs. In each epoch, one million training samples are randomly selected for training. We used a modified random sampler, where the weights of samples of class $C$ are set to $N_{C}^{-\lambda_{sampler}}$, with $N_C$ being the $C$ class frequency. We used the same value $\lambda_{sampler} = \frac{1}{2}$ as in the formula for selecting samples for the database. This approach softens class imbalances in the training set of each epoch, gives more focus on less frequent classes during training, and allows the model to learn traffic patterns from more diverse samples.

\subsubsection{ArcFace loss function}
In supervised metric learning, there are two main categories of loss functions used for training embedding models. The first consists of contrastive approaches, such as Contrastive~\cite{Chopra_2005} or Triplet~\cite{Schroff_2015} loss, which pull together embeddings of samples with the same label and push apart those with different labels. These methods operate within each mini-batch, using local sample-to-sample comparisons. The other group of loss functions includes softmax-based losses with class centers and margin modifications, such as ArcFace~\cite{Deng2019ArcFace} and CosFace~\cite{Wang_2018}. These methods introduce class-specific centers and enforce angular or cosine margins to better separate classes. Samples are pulled toward their class centers based on global sample-to-class comparisons. ArcFace, along with its sub-center variant~\cite{Deng2020Subcenter}, represents the current state-of-the-art. While originally developed for face recognition, ArcFace has proven effective across various tasks like image retrieval and fine-grained classification. The following paragraph contains a short overview of how the ArcFace loss works. For a more technical and formal description, please refer to the original ArcFace paper~\cite{Deng2019ArcFace}.

Training a neural network with the ArcFace loss involves adding an "ArcFace head" (see~\figref{fig:embedding-function}), which is detached and not used when the network is used for generating embeddings during validation and testing. The head contains a matrix of class centers, which are learnable during the training process. For computing the loss, both the embeddings and the class centers are normalized, which projects them onto a unit hypersphere. The ArcFace loss then calculates the angles between the flow embeddings and all the class centers. A fixed angular margin $m$ is added to the angle corresponding to the correct class, which improves class separation (a larger angular gap between neighboring class centers). The cosine of these angles is then computed and scaled by a parameter $s$, producing the logits. These scaled logits are then passed to a cross-entropy loss function to calculate the final loss. By incorporating the angular margin, the ArcFace loss encourages the model to cluster embeddings of the same class closer together while increasing the angular separation between different classes, resulting in more discriminative embeddings.

\subsubsection{Sub-center ArcFace with dynamic margins}
We experimented with an enhanced variant of ArcFace called sub-center ArcFace~\cite{Deng2020Subcenter}, which uses $K$ sub-centers per class instead of a single center. During training, samples are pulled toward the nearest positive sub-center. This loss is better suited for handling intra-class variations, which are common in TC tasks---for instance, due to different API endpoints hosted behind a single SNI domain. Moreover, we used an ArcFace variant with dynamic margins, as introduced in~\cite{Ha2020GoogleLandmark} for tasks with extreme class imbalance. Each class uses a different angular margin, calculated as $m_C = a * N_C^{-\lambda_{margin}} + b$, where $N_C$ is the $C$ class frequency, $a$ and $b$ define the minimum and maximum margins, and $\lambda_{margin} > 0$ controls the rate of change in the margin. We used $\lambda = \frac{1}{4}$, with $a$ and $b$ set to produce margins in the range [0.15, 0.25]. Less frequent classes, which require greater separation in the embedding space for accurate classification, are given larger margins (up to 0.25) to widen their decision boundaries, whereas more frequent classes are assigned smaller margins (down to 0.15).

\begin{table}[t]
\scriptsize
\centering
\setlength\extrarowheight{1pt}
\caption{An overview of the hyperparameter space.}
\label{tab:hyperparameters}
\begin{tabular}{c|m{6cm}}
    \multirow{3}{*}{\rotatebox[origin=c]{90}{DL arch.~~}} & Stem: size of embeddings (packet sizes, IPT), IPT bins edges, directions encoding, PLE initialization \\ \cline{2-2}
    & Convolutions: block architecture, number of blocks, per-block channels, strides, dropout rates \\ \cline{2-2}
    & Pooling operation, activation and normalization functions \\ \hline
    \multirow{4}{*}{\rotatebox[origin=c]{90}{Training loop~~~}} & Number of epochs, number of samples per epoch, semi-balanced sampling ($\lambda_{sampler}$) \\ \cline{2-2} % [-0.5cm]
    & ArcFace loss function: scale, margins, $K$ subcenters, $\lambda_{margin}$ for dynamic margins \\ \cline{2-2}
    & Initial LR, AdamW optimizer params, LR scheduling, number of warm-up iterations, weight decay, weight initialization \\ \cline{2-2}
    & KoLeo regularization strength  \\ \hline 
    \multirow{2}{*}{\rotatebox[origin=c]{90}{DB}} & Database size, semi-balanced database sampling ($\lambda_{db}$) \\ \cline{2-2}
    & Embedding size \\   
\end{tabular}
\end{table}

\subsubsection{Optimizer, LR scheduler, regularization, and implementation details}
\label{sec:optimizer}
The training loop was implemented in PyTorch. We used the AdamW optimizer with the default parameters, a batch size of 1024, and an initial learning rate of 0.0025. We used cosine learning rate decay, with a linear warm-up phase (from $\frac{0.0025}{3}$ to $0.0025$) for the first 150 iterations. Weight decay of $0.0017 (\approx10^{-2.75})$ was applied on all parameters except biases, BatchNorm affine parameters, packet size and IPT embedding matrices, and the GeM pooling $p$ parameter. All weights used PyTorch's default initialization, except for biases, which were set to zeros. We also use the KoLeo~\cite{sablayrolles2019spreadingvectorssimilaritysearch} regularization technique, which promotes a more uniform distribution of flow embeddings in the embedding space.

\subsection{Hyperparameter search}
The hyperparameter search was conducted on a single domain split (described in~\secref{sec:domain-preprocess}), and the best-found parameters were reused for other domain splits. Our goal was to identify a configuration with the highest macro-average recall on the validation domain set. Due to the large hyperparameter space, a full grid search was infeasible; instead, we optimized and fixed subsets of hyperparameters step by step. The most important hyperparameters are summarized in~\tabref{tab:hyperparameters}. In total, we used the MetaCentrum computing grid\footnote{\url{https://www.metacentrum.cz/en/}.} to run more than 4000 trials, each taking approximately two hours.

\section{Domain Recognition Results}
\label{sec:domain-recognition-results}
This section presents experimental results for the domain recognition task. First, we describe performance metrics, introduce a simple baseline approach, and present the results. We conclude this section with ablations analyzing the influence of various components and parameters in the experimental setup.

\subsection{Metrics}
\label{sec:metrics}
During validation and final testing, we perform database ranking to find the neighborhood of all query samples, as described in~\secref{sec:ranking}. The ranking relies on cosine similarity in the embedding space, where higher cosine similarity indicates more similar and closely related samples. To process a neighborhood into a domain prediction, we use three simple voting schemes: selecting the domain of the closest sample (top-1) or taking the majority domain among the three or five closest samples (maj-3, maj-5). In the case of ties (e.g., when all closest samples have different domains), the predicted domain is determined by the order of the neighboring samples. This approach is equivalent to using a k-NN classifier, where $k$ corresponds to the size of the neighborhood.

For each voting scheme, we compute classification accuracy and macro-average recall. Among these metrics, we consider macro-average recall to be more important because it reflects overall performance across all domains, regardless of their frequencies. To better understand performance differences between frequent and infrequent domains, we also calculate macro-average recall for quartiles of domains sorted by frequency. For example, Q1 recall represents the macro-average recall for the top 25\% most frequent domains, whereas Q4 recall corresponds to the bottom 25\% least frequent domains.

\subsection{Baseline definition}
\label{sec:input-space-baseline}
It is good practice to compare deep learning models against simple baseline methods to better understand the contribution of more complex solutions. To this end, we devised an input-space baseline that uses raw packet sequences as embeddings. The experimental setup for this baseline is identical to that of the $\Phi$ embedding function, but flows are represented using the first 10 packet sizes, directions, and scaled inter-packet times. Because inter-packet times are generally less informative than packet sizes, they are clipped to a maximum of one second and scaled by a factor of $\frac{1}{10}$ to reduce their relative influence in distance computation. The baseline uses L1 distance for ranking and the top-1 voting scheme for making predictions.

\subsection{Classification performance}
The final results for the domain recognition task were obtained as averages of 10 domain splits. Each domain split randomly divides the top 2000 domains into 1000 training domains, 500 validation domains, and 500 test domains. Additionally, for each domain split, we perform 10 repetitions, resulting in a total of 100 runs per reported value. Averaging over multiple domain splits ensures that the results are not biased for one specific set of domains. Results are presented in~\tabref{tab:main-results}, followed by a detailed discussion.

\begin{table}[t]
    \renewcommand{\arraystretch}{1.3}
    \begin{threeparttable}
    \centering
    \caption{Domain recognition results.}
    \label{tab:main-results}
    \setlength{\tabcolsep}{5pt}
    \begin{tabular}{|l|l|l|l|l|l|l|l|}
    \hline
       \multicolumn{2}{|c|}{\textbf{Method}} & \textbf{Acc} & \textbf{Recall} & \textbf{R-Q1} & \textbf{R-Q2} & \textbf{R-Q3} & \textbf{R-Q4} \\ \hline
        \multicolumn{2}{|c|}{\rule{0pt}{1.5em}Baseline} & \makecell[l]{71.44 \\[-1pt] \scriptsize{($\pm$4.81)}} & \makecell[l]{37.86 \\[-1pt] \scriptsize{($\pm$1.79)}} & \makecell[l]{51.91 \\[-1pt] \scriptsize{($\pm$1.26)}} & \makecell[l]{38.95 \\[-1pt] \scriptsize{($\pm$2.45)}} & \makecell[l]{32.77 \\[-1pt] \scriptsize{($\pm$2.72)}} & \makecell[l]{27.81 \\[-1pt] \scriptsize{($\pm$4.7)}} \\ \hline
        \multirow{3}{*}{\rotatebox[origin=c]{90}{Emb $\Phi$}} & \rule{0pt}{1.5em} Top-1 & \makecell[l]{94.83 \\[-1pt] \scriptsize{($\pm$0.98)}} & \makecell[l]{79.35 \\[-1pt] \scriptsize{($\pm$0.98)}} & \makecell[l]{89.63 \\[-1pt] \scriptsize{($\pm$0.97)}} & \makecell[l]{81.36 \\[-1pt] \scriptsize{($\pm$1.31)}} & \makecell[l]{75.99 \\[-1pt] \scriptsize{($\pm$1.54)}} & \makecell[l]{70.41 \\[-1pt] \scriptsize{($\pm$2.53)}} \\ \cline{2-8} 
        & Maj-3 & 95.3 & 79.01 & 90.15 & 81.3 & 75.35 & 69.26 \\ \cline{2-8}
        & Maj-5 & 95.54 & 78.37 & 90.44 & 80.94 & 74.38 & 67.73 \\ \hline 
    \end{tabular}
    \begin{tablenotes}
    \footnotesize
    \item Each value is an average over 100 runs (10 domain splits $\times$ 10 repetitions), except for the input-space baseline, which is averaged over 10 runs (10 domain splits $\times$ 1 repetition). Standard deviations for maj-3 and maj-5 are omitted to conserve space, as they do not differ from the top-1 values.
    \end{tablenotes}
    \end{threeparttable}
\end{table}

\subsubsection{Top-1 performance} 
The achieved top1-acc of 94.83\% and recall\footnote{All mentions of recall refer to top-1 recall, unless stated otherwise.} of 79.35\% are both surprising, and we consider them a success. This is remarkable given the challenging nature of our setup: the embedding function is evaluated on a set of domains disjoint from those used for training, and the task involves a large number of fine-grained classes. Measurements of recall across domain quartiles reveal interesting trends. For the most frequent Q1 domains, recall reaches 89.63\%. Between Q1 and Q2, there is a notable recall drop of around 8\%, and the subsequent gaps Q2 $\to$ Q3 and Q3 $\to$ Q4 are around 5\% each. It is evident that less frequent classes are much harder to recognize, a phenomenon that is well-known and understandable for a wide range of ML tasks. 

\subsubsection{Benefits of using a neighborhood with maj-3 and \mbox{maj-5}}
Considering a larger neighborhood of three or five samples introduces a trade-off between prioritizing Q1 domains and the rest. The Q1 recall improves for both maj-3 (90.15\%) and maj-5 (90.44\%) compared to top-1 (89.63\%). However, the Q2--Q4 recalls for both maj-3 and maj-5 are lower than for top-1. This decrease can be attributed to sparse embeddings of less frequent domains, which have fewer database samples and, therefore, lack sufficient representation to "win" in the maj-3 and maj-5 voting schemes. Overall, using a larger neighborhood proves advantageous for the top 25\% most frequent domains, while top-1 works better for the rest. The improved performance on the most frequent domains also explains that the classification accuracies of maj-3 (95.3\%) and maj-5 (95.54\%) exceed that of top-1 (94.83\%), as frequent classes have a significant impact on micro-averaged metrics.

\subsubsection{Baseline performance}
The input-space baseline was evaluated using the same experimental setup as the proposed $\Phi$ embedding function, enabling a direct performance comparison. The results show a significant 23.39\% improvement in top1-acc of the proposed $\Phi$ embedding function over the baseline and even bigger improvements in Q1--Q4 recalls. This demonstrates that the baseline method is inadequate for addressing the domain recognition task within the given setup. However, our transfer learning experiments revealed that the input-space baseline can match SOTA performance on other TC datasets; see~\ref{sec:input-space-transfer-results} for more results and related discussion.

\subsection{Ablations}
The purpose of ablation studies is to investigate a system's performance by removing or modifying certain components to gain a better understanding of their contributions to the overall system.
The following sections examine the role of certain hyperparameters and their effects on classification performance and ranking speed. All ablation experiments were conducted using one specific domain split (identical to that used in the hyperparameter search), and we report average results from 10 runs per configuration, if not stated otherwise.

\begin{table}[b]
    \centering
    \footnotesize
    \setlength\extrarowheight{1pt}
    \caption{Comparison of approaches for encoding packet features.}
    \label{tab:packet-embedding}
    \begin{tabular}{|l|l|l|l|l|l|l|}
    \hline
        \textbf{Method} & \textbf{Top1-Acc} & \textbf{Recall} & \textbf{R-Q1} & \textbf{R-Q2} & \textbf{R-Q3} & \textbf{R-Q4} \\ \hline
        Emb+PLE & 95.75 & 79.29 & 90.38 & 79.59 & 74 & 73.18 \\ \hline
        PLE  & 95.67 & 78.49 & 90.19 & 78.64 & 73.03 & 72.11 \\ \hline
        Scalars & 95.35 & 76.23 & 89.46 & 76.89 & 69.95 & 68.61 \\ \hline
        Emb+Rand & 95.02 & 75.73 & 88.8 & 76.1 & 70.25 & 67.75 \\ \hline
    \end{tabular}
\end{table}

\subsubsection{Packet features - direct scalar values, PLE encoding, or learnable Embedding layer with PLE initialization}
\label{sec:ple}
We evaluated the impact of our packet embedding scheme, referred to as Emb+PLE and detailed in~\secref{sec:stem}, and present the results in \tabref{tab:packet-embedding}. Most related works use direct scalar values, as did our previous architectures. In terms of recall, our embedding scheme (79.29\%) shows a 3.06\% improvement over the scalar approach (76.23\%). The benefits are most prominent for less frequent domains, with a Q4 recall improvement of 4.57\%. We also tested a variant denoted as PLE, which uses the initial embeddings created with PLE encoding without further optimization (i.e., \textit{Embedding} weights are frozen). The results show PLE encoding accounts for most benefits, while the trainable \textit{Embedding} layer adds a smaller incremental improvement. The gains in top1-acc are more modest, with Emb+PLE providing a 0.4\% improvement over scalars.

Furthermore, we evaluated a variant denoted as Emb+Rand, which is similar to the Emb+PLE scheme but uses random initialization for the embedding matrices. It is likely that related works utilizing packet embeddings~\cite{Nascita2021FirstEmbeddings, Nascita2023Embeddings} employ this variant, as no initialization method is specified and random initialization is the common default. \tabref{tab:packet-embedding} shows that Emb+Rand underperforms, further highlighting the strength of PLE initialization, which produces embeddings with inherent ordering and thereby facilitates the training process.

\subsubsection{Training sampler - the impact of the balancing parameter}
\label{sec:ablation-training-sampler}
In each training epoch, one million network flows are sampled using a semi-balanced random sampler, where the $\lambda_{sampler}$ parameter controls the balancing strength (see details in~\secref{sec:training-sampler}). When $\lambda_{sampler} = 0$, all samples are assigned equal weight, and the original imbalance is preserved (no balancing). When $\lambda_{sampler} = 1$, the sampler creates a per-domain distribution that is as uniform as possible (perfectly uniform distribution is unattainable as we use sampling without replacement).

\begin{figure}[t!]
    \centering
    \includegraphics[width=0.98\columnwidth]{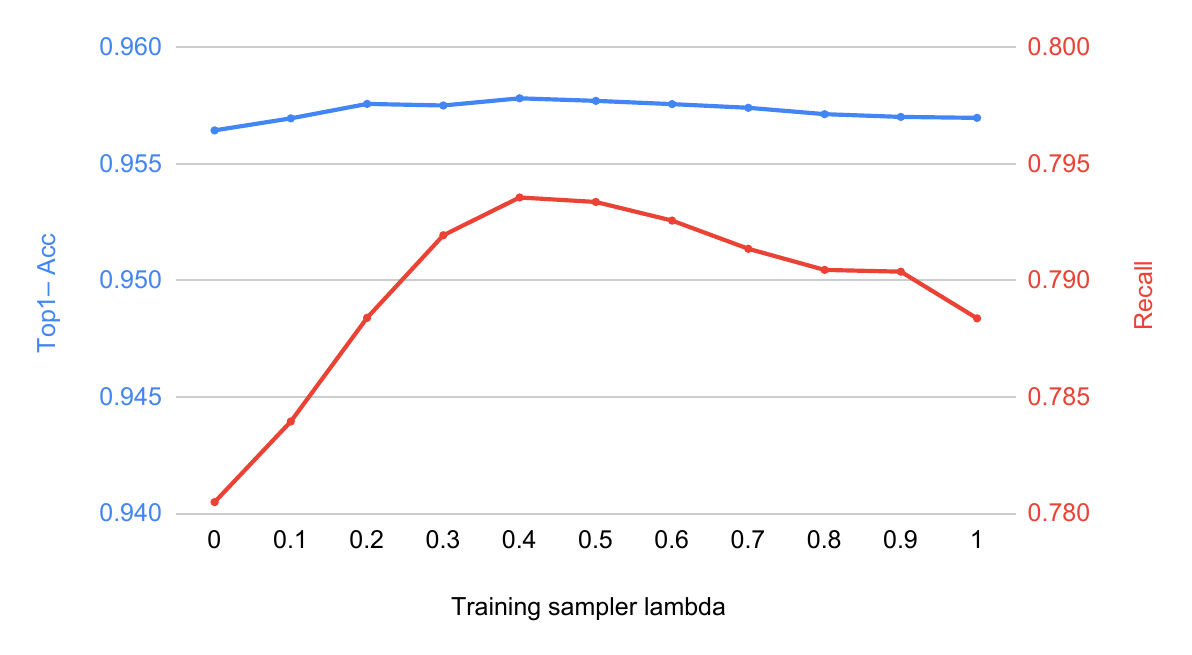}
    \caption{The impact of the $\lambda_{sampler}$ balancing parameter of training sampler. Both vertical axes of top1-acc and recall have a range of 2\% to make the shapes of the lines comparable.}
    \label{fig:sampler-lambda}
\end{figure}

\begin{table*}[t]
    \centering
    \setlength\extrarowheight{1pt}
    \caption{The combined effect of both database balancing  $\lambda_{db}$ and $\lambda_{sampler}$.}
    \label{tab:db-sampler-lambdas}
    \begin{tabular}{|cc|l|l|l|l|l|l|}
        \hline
         $\lambda_{db}$ & $\lambda_{sampler}$ & \textbf{Top1-Acc} & \textbf{Recall} & \textbf{R-Q1} & \textbf{R-Q2} & \textbf{R-Q3} & \textbf{R-Q4} \\ \hline
        0.0 & 0.0 & 96.15                      & 70.6 & 88.52 & 71.2 & 62.96 & 59.71                              \\ \hline
        0.0 & 0.5 & 96.26 \scriptsize{(+0.11)} & 72.29 \scriptsize{(+1.69)} & 88.95 \scriptsize{(+0.43)} & 72.70 \scriptsize{(+1.5)} & 64.75 \scriptsize{(+1.79)} & 62.74 \scriptsize{(+3.03)} \\ \hline
        0.5 & 0.0 & 95.62 \scriptsize{(-0.53)} & 78.00 \scriptsize{(+7.4)} & 90.06 \scriptsize{(+1.54)} & 78.42 \scriptsize{(+7.22)} & 72.57 \scriptsize{(+9.61)} & 70.95 \scriptsize{(+11.24)} \\ \hline
        0.5 & 0.5 & 95.75 \scriptsize{(-0.4)}  & 79.29 \scriptsize{(+8.69)} & 90.38 \scriptsize{(+1.86)} & 79.59 \scriptsize{(+8.39)} & 74.00 \scriptsize{(+11.04)} & 73.18 \scriptsize{(+13.47)} \\ \hline
    \end{tabular}
\end{table*}

Our expectation was that changing $\lambda_{sampler}$ would provide a trade-off between focusing on more frequent with $\lambda_{sampler} = 0$ (higher top1-acc) or less frequent domains with $\lambda_{sampler} = 1$ (higher recall). However, it turned out that some degree of balancing has a positive impact even for the top1-acc metric. We ended up choosing $\lambda_{sampler} = \frac{1}{2}$, meaning the final weight formula is $\frac{1}{\sqrt{N_C}}$. Compared to a standard random sampler without balancing (which corresponds to $\lambda_{sampler} = 0$), this brings a 1.29\% improvement in recall and a minuscule improvement of 0.13\% in top1-acc. The impact on both metrics is showcased in~\figref{fig:sampler-lambda}.

\subsubsection{Database - the impact of the balancing parameter}
\label{sec:ablation-db-sampler}
\begin{figure}[t!]
    \centering
    \includegraphics[width=0.98\columnwidth]{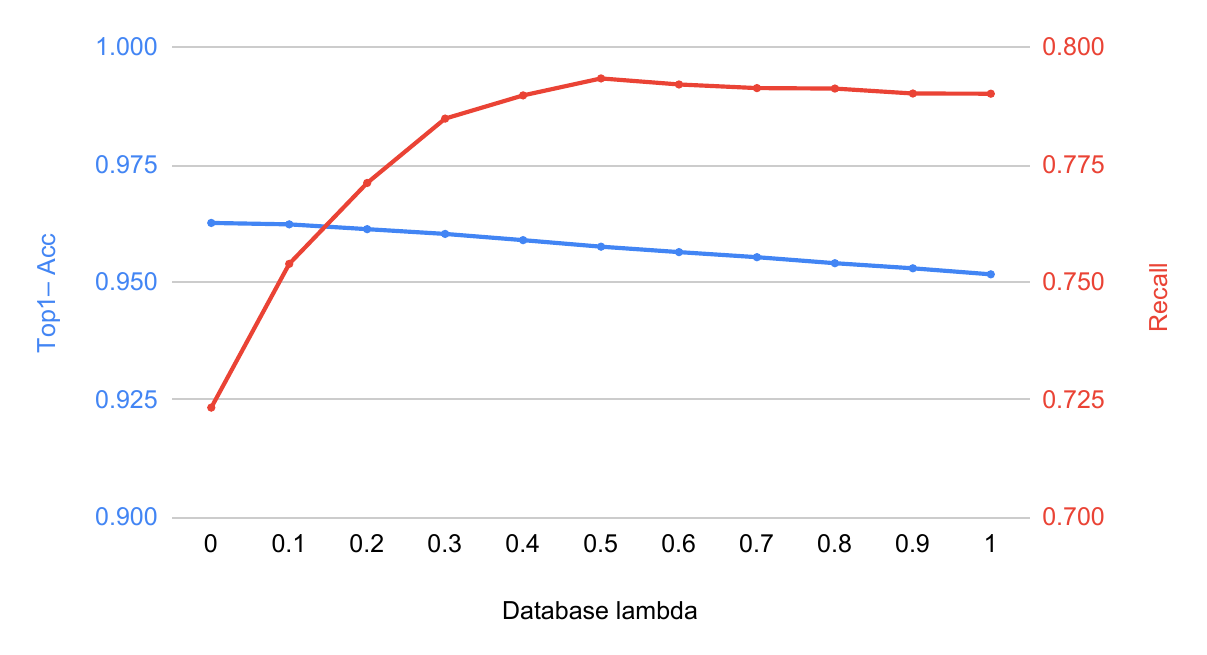}
    \caption{The impact of the $\lambda_{db}$ balancing parameter. Both vertical axes of top1-acc and recall have a range of 10\% to make the shapes of the lines comparable.}
    \label{fig:db-lambda}
\end{figure}

As described in~\secref{sec:database-preparation}, we also perform semi-balanced sampling for building the database. The weight formula is the same as for the epoch training sampler. When $\lambda_{db} = 0$, the original imbalance is preserved in the database. When $\lambda_{db} = 1$, the database has the per-domain distribution as uniform as possible (perfectly uniform distribution is unattainable as we use sampling without replacement). The graph investigating the impact of the $\lambda_{db}$ parameter in~\figref{fig:db-lambda} shows a clear trade-off between decreasing top1-acc and increasing recall when $\lambda_{db}$ moves from 0 to 1. We ended up choosing $\lambda_{db} = \frac{1}{2}$, which, compared to no balancing, brings a 7\% improvement in recall at the expense of a 0.51\% decrease in top1-acc. We believe this trade-off is worthwhile in most scenarios.

\paragraph*{Combined effect}
We also examined the combined effect of $\lambda_{sampler}$ and $\lambda_{db}$. The results, presented in~\tabref{tab:db-sampler-lambdas}, highlight the performance gains compared to the case where neither the database nor the training set is balanced ($\lambda_{sampler} = 0 $ and $\lambda_{db} = 0$).\footnote{\figref{fig:sampler-lambda} shows the change between third and fourth rows of~\tabref{tab:db-sampler-lambdas} with $\lambda_{db} = \frac{1}{2}$ fixed, while \figref{fig:db-lambda} shows the change between second and fourth rows with $\lambda_{sampler} = \frac{1}{2}$ fixed.} As expected, balancing has the greatest impact on Q4 recall, for which it provides a remarkable 13.47\% improvement. The gains for Q3 (+11.04\%) and Q4 (+8.39\%) recalls are also impressive, especially given that the "cost" is merely a 0.4\% decrease in top1-acc. Furthermore, the results indicate that database balancing and training set balancing operate independently, as their combined effect is approximately the sum of their individual contributions.

\subsubsection{Database - how does the number of unique domains affect performance?}
This section examines the sensitivity of the domain recognition approach to the number of unique domains we want to recognize. Previous work~\cite{Yang2021} demonstrated that TC tasks often become trivial when the number of classes is small. To explore this, we tested our approach with the number of unique domains ranging from 100 to 1000.

\begin{figure}[t!]
    \centering
    \includegraphics[width=1\columnwidth]{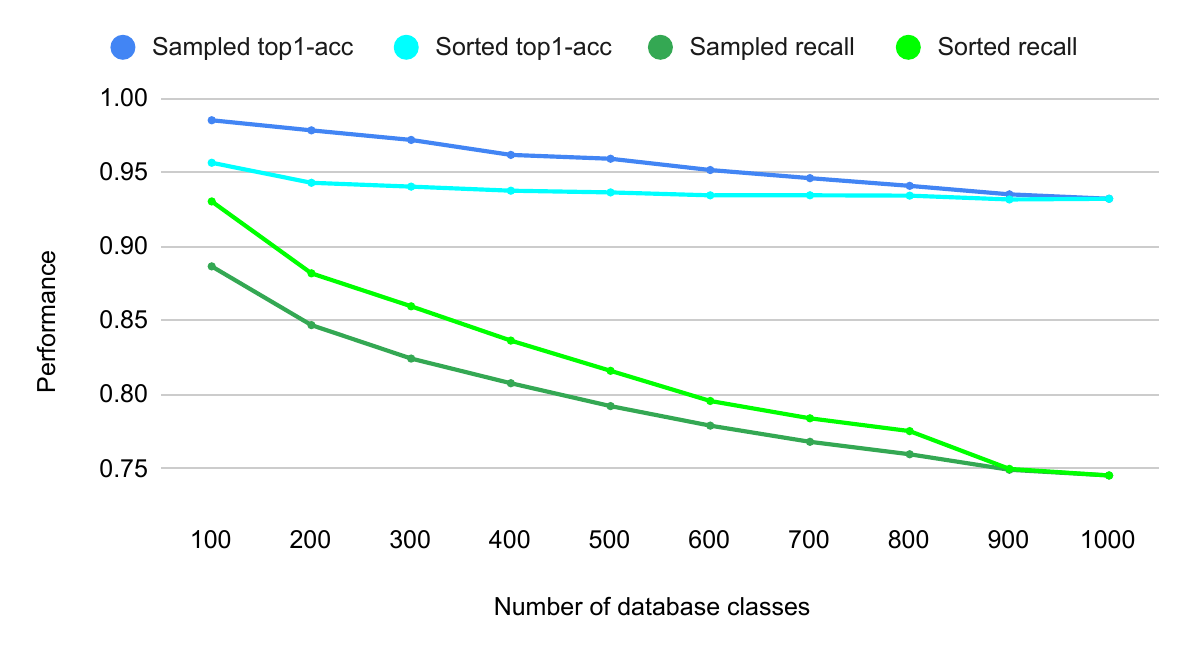}
    \caption{Sensitivity of the proposed domain recognition approach to the number of unique domains.}
    \label{fig:db-classes}
\end{figure}

To obtain up to 1000 domains, we first merge the validation and test domain sets.\footnote{We acknowledge that reusing validation domains for testing deviates from our defined evaluation protocol; however, we did so only in this experiment to demonstrate performance across a wider range of classes.} Then, we randomly select the desired number of domains, choose all samples of those domains, and split them into query and database parts. A database is created using the $\lambda_{db}$ semi-balanced sampling. We measure the performance of a single trained model, which is reused for all domain counts and repetitions. For each domain count, we repeat this procedure 50 times and report the average. For the maximum of 1000 domains, we use all available validation and test domains. Additionally, we compare the described random domain sampling with a sorted approach, where we select the $N$ (100, 200, \dots, 1000) most frequent domains in each repetition. The results are presented in~\figref{fig:db-classes}. The top1-acc for random domain sampling ranges from 98.52\% for 100 domains to 93.22\% for 1000 domains, while the recall ranges from 88.65\% to 74.54\%. In contrast, the recall for the top 100 most frequent domains is as high as 93.04\%, which is notable considering these 100 domains cover 85\% of all dataset samples.

\begin{figure}[t!]
    \centering
    \includegraphics[width=1\columnwidth]{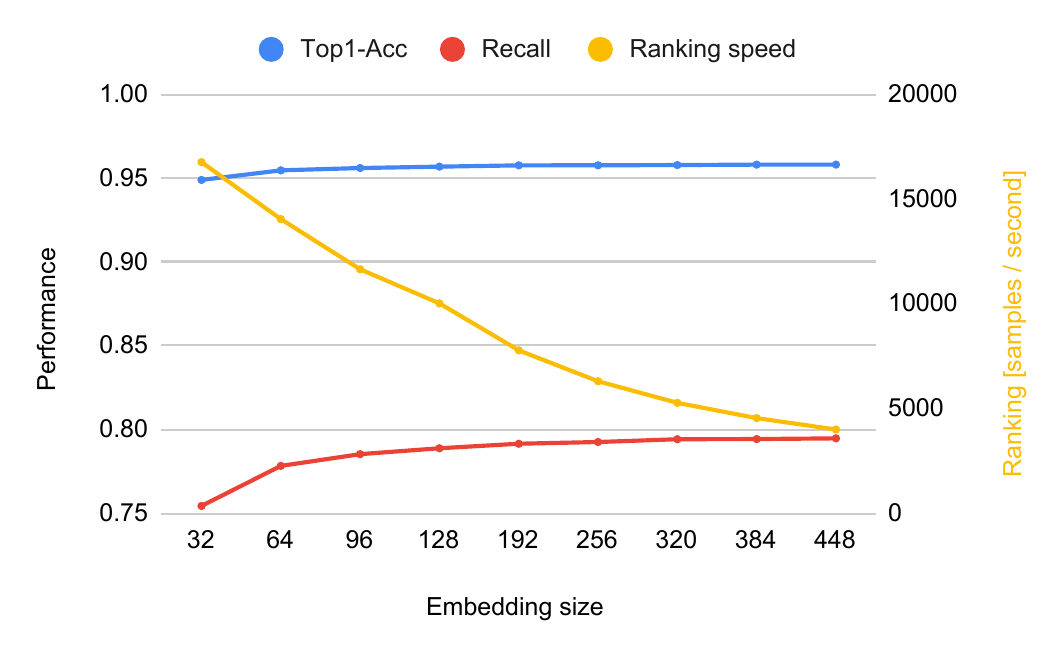}
    \caption{The impact of the flow embedding size on domain recognition accuracy and ranking speed.}
    \label{fig:embedding-size-ranking-speed}
\end{figure}

An interesting difference is observed between the domain selection methods: top1-acc is higher on random subsets of domains, while recall is better when using the top $N$ domains. This is because when domains are sampled, some less frequent and harder-to-recognize domains are included. Since recall is macro-averaged, these harder domains have a larger impact on the overall metric. In contrast, for micro-averaged top1-acc, the inclusion of less frequent domains has minimal effect on the overall metric. Moreover, top1-acc is higher with sampled domains because there are fewer misclassifications in the region of the most frequent domains. Among these domains, there are a lot of similar ones that are prone to mismatch, and thus, top1-acc is increased when some of those are not selected in the given repetition.

\subsubsection{Embedding size - ranking speed vs recall trade-off}
\label{sec:ablation-ranking-speed}
The flow embedding size is a key parameter that influences the performance of the proposed domain recognition approach. Larger embeddings improve recognition accuracy but come at the cost of slower ranking speeds. To explore this trade-off, we ran experiments with embedding sizes ranging from 32 to 448. As with other ablations, we performed 10 repetitions per embedding size and report the average metrics. The results, summarized in~\figref{fig:embedding-size-ranking-speed}, show that reducing the embedding size has a limited impact on top1-acc (94.88\% for size 32, 95.79\% for size 448). However, recall is more sensitive, decreasing from 79.46\% (size 448) to 75.43\% (size 32).

We observed a clear inverse relationship between ranking speed and embedding size: the smallest embeddings achieved speeds of around 16.5k flows/s, while the largest slowed the ranking to 4k flows/s. In contrast, the speed of creating the embeddings with the neural network remained stable at around 33k flows/s, regardless of the used embedding size. Both tasks---creating embeddings and faiss\footnote{The \textit{faiss} library, which we use for database ranking, supports GPU indexes that offer a 5$\times$ - 10$\times$ performance boost compared to CPU implementations. \url{https://github.com/facebookresearch/faiss/wiki/Faiss-on-the-GPU}.} ranking---were performed on an Nvidia Tesla T4 16GB GPU. Further performance-related discussion is provided in the final chapter.

\begin{table}[h]
    \footnotesize
    \centering
    \setlength\extrarowheight{1pt}
    \caption{KoLeo regularization impact.}
    \label{tab:koleo}
    \begin{tabular}{|l|l|l|l|l|l|}
    \hline
        \textbf{KoLeo} & \textbf{Best epoch} & \textbf{Top1-Acc} & \textbf{Recall}\\ \hline
        Yes ($\lambda = 1$) & 24.3 & 95.75 & 79.29  \\ \hline
        No ($\lambda = 0$) & 22.6 & 95.68 & 78.97 \\ \hline
    \end{tabular}
\end{table}

\subsubsection{KoLeo regularization}
We investigated the impact of KoLeo regularization, which uses a parameter $\lambda$ to control its strength. Although KoLeo was originally designed~\cite{sablayrolles2019spreadingvectorssimilaritysearch} to improve embedding discretization---a step we do not perform---we observed that without this regularization, training diverges before completing the 30 training epochs. \tabref{tab:koleo} compares the results with ($\lambda=1$) and without ($\lambda=0$) KoLeo. When KoLeo was used, the best validation performance was achieved later in training, indicating the neural network continued improving over more epochs. The later peak in validation performance suggests that KoLeo contributes to a more stable training process and enhances resistance to overfitting. However, the performance gains are modest: a 0.32\% improvement in recall and a 0.07\% in top1-acc. 

\section{Transfer to downstream traffic classification tasks}
\label{sec:transfer-learning}
To assess how well the proposed $\Phi$ embedding function pretrained on domain recognition generalizes, we evaluated it on ten established TC tasks and compared its performance to published SOTA results. We tested three transfer methods for each task: a k-NN classifier with fixed embeddings, a linear classifier with fixed embeddings (known as linear probing), and a linear classifier with fine-tuning of the embedding model. For each task, we also trained the same model from scratch in order to isolate the performance gains of the transfer learning methods. Furthermore, we measured the performance of the input-space baseline, which was defined in~\secref{sec:input-space-baseline}. Overall, we report six measurements for each downstream TC task: SOTA, input-space baseline, training from scratch, k-NN transfer method, linear probing, and embeddings fine-tuning. The following sections describe transfer methods, provide an overview of the used datasets, and discuss results.

\subsection{Transfer methods}
\label{sec:transfer-methods}
For all transfer methods, we used ``intermediate embeddings'' of the $\Phi$ embedding function, taken after the global pooling and before the feature refinement block (see Figure \ref{fig:backbone-model}). Features from deeper layers capture more general patterns and consistently offered better results for all tested methods.

\subsubsection{k-NN transfer with fixed embeddings}
For the k-NN transfer method, we created embeddings for all samples of the given downstream task, trained a k-NN classifier on embeddings of the training set, and evaluated it on the test set. Predictions were based on the label of the closest training sample in the embedding space, and we did not use the $\lambda_{db}$ semi-balancing technique. 

\subsubsection{Linear probing with fixed embeddings}
For linear probing, the setup was the same as for the k-NN method: we generated embeddings for the downstream task and trained a linear classifier on them. For smaller datasets (all except CESNET-TLS22 and AppClassNet), we used an exact solver from scikit-learn to fit the classifier, while for the large datasets, we used a single PyTorch Linear layer and implemented a simple training loop for it. Using the exact solver provided performance gains of around 1\%, but it did not converge in a reasonable time on large datasets.

\subsubsection{Fine-tuned model}
The fine-tuning method uses a linear classification head and allows the entire \texttt{30pktTCNET} backbone to be updated during training on the downstream task. Fine-tuning was performed over 50 epochs using the AdamW optimizer and a cosine learning rate schedule with a warm-up phase. All BatchNorm layers were kept in evaluation mode to preserve the batch statistics learned during pretraining. For each downstream task, we conducted a hyperparameter search over the learning rate, warm-up length, batch size, dropout rate of the classification head, and the pooling operation (allowing the original GeM pooling to be replaced with either max or average pooling). Hyperparameters were selected based on validation performance on the first data split.

If not addressed during fine-tuning, a neural network starts to forget its original capabilities and the general knowledge acquired during the pretraining task. To mitigate this, we implemented three fine-tuning techniques. \textit{(a)} We use lower learning rates for deeper layers~\cite{Dong2022_LRdecay}, which extract low-level traffic patterns analogous to edges and shapes in image processing. Along with other hyperparameters, we optimize a multiplicative factor, $LR\_mult \in (0, 1]$, which scales the learning rate of the backbone's residual blocks based on their depth, using the formula: $LR_{block} = LR \times LR\_mult^{depth}$. \textit{(b)} We applied L2 Starting Point (L2SP) regularization~\cite{Li2018_L2SP}, a technique designed to mitigate catastrophic forgetting (i.e., the loss of knowledge from previous tasks). It adds a regularization term based on the L2 distance between the current model weights and the original pretrained weights (the starting point). This discourages large deviations from the pretrained weights, helps retain prior knowledge, and reduces overfitting on downstream tasks with limited training data. \textit{(c)} A related but more recent method, called L2 Distance in Feature Space (LDIFS)~\cite{Mukhoti2024_LDIFS}, adds a regularization term based on the L2 distance between embeddings of the current batch created with the original (pretrained) model and the updated model. This encourages alignment between the original and fine-tuned feature spaces. While Mukhoti et al.~\cite{Mukhoti2024_LDIFS} applied LDIFS to preserve intermediate features, we found that focusing on final features was sufficient for our use case. Both L2SP and LDIFS are controlled with regularization strength hyperparameters, where setting the value to zero disables the corresponding regularization. For downstream tasks with larger training sets (CESNET-TLS22, AppClassNet, MIRAGE19), neither regularization term was necessary. Three tasks benefited from a combination of both techniques, while the remaining tasks achieved the best fine-tuned performance using LDIFS alone.

\subsection{Code and pretrained model availability}
We published the embedding model and its pretrained weights in the CESNET Models framework~\cite{Luxemburk2024_CesnetModels} under the name \texttt{30pktTCNET\_256}. The architecture code is available on GitHub.\footnote{\url{https://github.com/CESNET/cesnet-models/blob/main/cesnet_models/architectures/multimodal_cesnet_enhanced.py}.} To select the best model (one specific set of weights), we chose the one with the highest sum of validation and test recalls on the domain recognition task, considering all training runs with the final hyperparameter configuration. Prior to transfer to downstream tasks, we made a single modification to the model, targeting the packet size embedding technique implemented in the model stem.\footnote{In the model stem, the \textit{Embedding} layer generates a learned vector representation for each packet size. However, some packet sizes (e.g., 1453--1471, 1473--1500) are never observed during training, leaving their representations untrained. To address this, we assign them the representation of the nearest observed packet sizes, except for packet sizes 1--19, which are given the representation of zero packet size.} Thanks to the model's publication and the availability of open-source tools that provide access to the datasets, all transfer learning experiments presented in this section are reproducible. To support this, we have published our transfer learning codebase\footnote{\url{https://github.com/CESNET/tc-transfer}.}, which enables replication of the results presented in Tables~\ref{tab:transfer-methods-results} and~\ref{tab:cross-eval-results}.

\subsection{Datasets}
\label{sec:transfer-datasets}
To evaluate transfer learning methods, we used seven additional datasets: ISCXVPN2016~\cite{dataset_ISCXVPN2016}, MIRAGE19~\cite{dataset_MIRAGE19}, MIRAGE22~\cite{dataset_MIRAGE22}, UTMOBILENET21~\cite{dataset_UTMOBILENET21}, UCDAVIS19~\cite{dataset_UCDAVIS19}, CESNET-TLS22~\cite{Luxemburk2023TLS}, and AppClassNet~\cite{dataset_AppClassNet}, covering ten classification tasks in total. To streamline dataset handling, we used the \textit{tcbench}\footnote{\url{https://github.com/tcbenchstack/tcbench}.} framework, which provides four datasets: MIRAGE19, MIRAGE22, UTMOBILENET21, and UCDAVIS19. These datasets are cleaned, pre-filtered when necessary, and include prepared train/validation/test splits. Overall, \textit{tcbench} significantly simplified cross-dataset evaluation and saved us considerable time. CESNET-TLS22 was obtained via the CESNET DataZoo toolset~\cite{Luxemburk2023Datazoo}, AppClassNet was downloaded from its official Figshare repository~\cite{dataset_AppClassNet_figshare}, and ISCXVPN2016 was kindly provided by Nascita et al.~\cite{Nascita2023Embeddings}. The following sections introduce each dataset and explain how it was used. We also provide related work used for SOTA comparison, with additional details about SOTA performance available in Appendix~\ref{ap:sota-comparison}. For all datasets, train/validation/test splits were used as follows: the training set was used to fit the k-NN classifier, perform linear probing, or fine-tune the model---depending on the selected transfer method. The validation set was used for hyperparameter search, and we report the average performance across all test splits.

\begin{table*}[!ht]
    \centering
    \setlength\extrarowheight{1.2pt}
    \caption{Downstream performance of the three transfer methods. Deltas are computed relative to training from scratch on each downstream task.}
    \label{tab:transfer-methods-results}
\begin{threeparttable}
\begin{NiceTabular}{|l|c|c|l|lr|lr|lr|}
\hline
\multicolumn{3}{|c|}{\textbf{Dataset}} &
\multicolumn{7}{c|}{\textbf{Downstream Task Performance [\%]}}
\\ \hline
%-----------------------------------------------------------
\textbf{Name} & 
\textbf{Subtask} &
\textbf{Classes} &
\multicolumn{1}{c|}{\textbf{From Scratch}} &
\multicolumn{2}{c|}{\textbf{Linear Probing}} &
\multicolumn{2}{c|}{\textbf{k-NN Transfer}} &
\multicolumn{2}{c|}{\textbf{Fine-Tuned Model}} 
\\ \hline
%-----------------------------------------------------------
ISCXVPN2016&
Application& 
15&
75.71 \scriptsize{($\pm$0.62)} & 
\cellcolor[HTML]{FFFFFF} 72.81 & 
\cellcolor[HTML]{FFE5E5} \textit{($\Delta$\:\:\:\:\:$-$2.9)} & 
\cellcolor[HTML]{FFFFFF} 73.25 & 
\cellcolor[HTML]{FFE5E5} \textit{($\Delta$\:$-$2.46)} & 
\cellcolor[HTML]{FFFFFF} 77.44 \scriptsize{($\pm$0.83)} & 
\cellcolor[HTML]{DDF2E8} \textit{($\Delta$\:\:\:\,1.73)} 
\\ \hline
%-----------------------------------------------------------
ISCXVPN2016&
Traffic type& 
6&
78.35 \scriptsize{($\pm$0.95)} & 
\cellcolor[HTML]{FFFFFF} 73.01 & 
\cellcolor[HTML]{FFB8B8} \textit{($\Delta$\:\:\:$-$5.34)} & 
\cellcolor[HTML]{FFFFFF} 75.57 & 
\cellcolor[HTML]{FFE5E5} \textit{($\Delta$\:$-$2.78)} &
\cellcolor[HTML]{FFFFFF} 80.19 \scriptsize{($\pm$0.92)} & 
\cellcolor[HTML]{DDF2E8} \textit{($\Delta$\:\:\:\,1.84)} 
\\ \hline
%-----------------------------------------------------------
ISCXVPN2016&
Encapsulation& 
2&
93.96 \scriptsize{($\pm$0.45)} & 
\cellcolor[HTML]{FFFFFF} 88.03 & 
\cellcolor[HTML]{FFB8B8} \textit{($\Delta$\:\:\:$-$5.93)} & 
\cellcolor[HTML]{FFFFFF} 91.83 & 
\cellcolor[HTML]{FFE5E5} \textit{($\Delta$\:$-$2.13)} &
\cellcolor[HTML]{FFFFFF} 94.37 \scriptsize{($\pm$0.45)} & 
\cellcolor[HTML]{DDF2E8} \textit{($\Delta$\:\:\:\,0.41)}  
\\ \hline
%-----------------------------------------------------------
MIRAGE19 &
&
20&
82.49 \scriptsize{($\pm$0.36)} &
\cellcolor[HTML]{FFFFFF} 72.51 & 
\cellcolor[HTML]{FFB8B8} \textit{($\Delta$\:\:\:$-$9.98)} & 
\cellcolor[HTML]{FFFFFF} 84.75 &
\cellcolor[HTML]{DDF2E8} \textit{($\Delta$\:\:\:\:\,2.26)} &
\cellcolor[HTML]{FFFFFF} 85.31 \scriptsize{($\pm$0.35)} &
\cellcolor[HTML]{DDF2E8} \textit{($\Delta$\:\:\:\,2.82)} 
\\ \hline
%-----------------------------------------------------------
MIRAGE22 &
& 
9&
98.48 \scriptsize{($\pm$0.15)} &
\cellcolor[HTML]{FFFFFF} 95.89 & 
\cellcolor[HTML]{FFE5E5} \textit{($\Delta$\:\:\:$-$2.59)} &
\cellcolor[HTML]{FFFFFF} 98.19 &
\cellcolor[HTML]{FFE5E5} \textit{($\Delta$\:$-$0.29)} &
\cellcolor[HTML]{FFFFFF} 98.67 \scriptsize{($\pm$0.27)} &
\cellcolor[HTML]{DDF2E8} \textit{($\Delta$\:\:\:\,0.19)}
\\ \hline
%-----------------------------------------------------------
UTMOBILENET21 &
&
17&
85.3\:\:\:\:\scriptsize{($\pm$0.94)} & 
\cellcolor[HTML]{FFFFFF} 86.43 & 
\cellcolor[HTML]{DDF2E8} \textit{($\Delta$\:\:\:\:\:\:\,1.13)} & 
\cellcolor[HTML]{FFFFFF} 86.85 & 
\cellcolor[HTML]{DDF2E8} \textit{($\Delta$\:\:\:\:\,1.55)} & 
\cellcolor[HTML]{FFFFFF} 88.9\:\:\:\:\scriptsize{($\pm$0.75)} & 
\cellcolor[HTML]{57BB8A} \textit{($\Delta$\:\:\:\:\:\:3.6)}  
\\ \hline
%-----------------------------------------------------------
UCDAVIS19&
\textit{Script} test& 
5&
98.73 \scriptsize{($\pm$0.2)} & 
\cellcolor[HTML]{FFFFFF} 98 & 
\cellcolor[HTML]{FFE5E5} \textit{($\Delta$\:\:\:$-$0.73)} & 
\cellcolor[HTML]{FFFFFF} 100 & 
\cellcolor[HTML]{DDF2E8} \textit{($\Delta$\:\:\:\:\,1.27)} & 
\cellcolor[HTML]{FFFFFF} 98.13 \scriptsize{($\pm$0.27)} & 
\cellcolor[HTML]{FFE5E5} \textit{($\Delta$\:\:\,$-$0.6)} 
\\ \hline
%-----------------------------------------------------------
UCDAVIS19 &
\textit{Human} test&
5&
78.07 \scriptsize{($\pm$2.14)} &
\cellcolor[HTML]{FFFFFF} 87.47 & 
\cellcolor[HTML]{57BB8A} \textit{($\Delta$\:\:\:\:\:\:\:\:\:9.4)} & 
\cellcolor[HTML]{FFFFFF} 84.82 &
\cellcolor[HTML]{57BB8A} \textit{($\Delta$\:\:\:\:\,6.75)} &
\cellcolor[HTML]{FFFFFF} 89.64 \scriptsize{($\pm$1.1)} &
\cellcolor[HTML]{57BB8A} \textit{($\Delta$\:\:11.57)}
\\ \hline
%-----------------------------------------------------------
CESNET-TLS22  &
& 
191&
98.03 \scriptsize{($\pm$0.03)} &
\cellcolor[HTML]{FFFFFF} 85.65 & 
\cellcolor[HTML]{FFB8B8} \textit{($\Delta$\:$-$12.38)} & 
\cellcolor[HTML]{FFFFFF} 96.44 &
\cellcolor[HTML]{FFE5E5} \textit{($\Delta$\:$-$1.59)} &
\cellcolor[HTML]{FFFFFF} 98.25 \scriptsize{($\pm$0.02)} &
\cellcolor[HTML]{DDF2E8} \textit{($\Delta$\:\:\:\,\,0.22)}
\\ \hline 
%-----------------------------------------------------------
AppClassNet  &
& 
200&
92.42 \scriptsize{($\pm$0.02)} &
\cellcolor[HTML]{FFFFFF} 62.15 & 
\cellcolor[HTML]{FFB8B8} \textit{($\Delta$\:$-$30.27)} & 
\cellcolor[HTML]{FFFFFF} 81.12 &
\cellcolor[HTML]{FFB8B8} \textit{($\Delta$\:$-$11.3)} &
\cellcolor[HTML]{FFFFFF} 92.11 \scriptsize{($\pm$0.03)} &
\cellcolor[HTML]{FFE5E5} \textit{($\Delta$\:$-$0.31)} 
\\ \hline 

\end{NiceTabular}

\begin{tablenotes}
\leftskip=0.3cm

\item Results are averages across ten splits. For \textit{tcbench} datasets with five prepared splits, we used each split twice. Standard deviations of linear probing \\\;and k-NN transfer are omitted to conserve space, but they do not differ considerably from the values shown. All results are classification accuracies.

\end{tablenotes}
\end{threeparttable}

\end{table*}

\subsubsection*{ISCXVPN2016}
A lab-generated traffic dataset that covers three classification tasks: encapsulation (VPN vs. nonVPN), traffic types, and applications. We used the preprocessed version provided by Nascita et al.~\cite{Nascita2023Embeddings}, where broadcast flows and other noisy data were filtered out. We divided the dataset into ten stratified 60/20/20 train/validation/test splits. For SOTA comparison, we rely on the results reported by Nascita et al., who evaluated several models---some leveraging payload (e.g., the multi-modal network \texttt{DISTILLER-Embeddings}~\cite{Nascita2023Embeddings}), and others that do not (e.g., the 1D CNN over packet sequences by Rezaei et al.\cite{Rezaei2020Multitask}). Since we focus on TC solutions that do not use payload data, we primarily compare against models operating under this same constraint. 

\subsubsection*{MIRAGE19}
A well-known mobile traffic dataset from the MIRAGE dataset series. The traffic of this dataset is based on real users interactions with 20 Android applications. A private version that contains 40 applications exists but is not part of \textit{tcbench}. Aceto et al.~\cite{dataset_MIRAGE19} published this dataset in 2019, providing JSON files containing traffic capture experiments. The authors of \textit{tcbench} processed the JSON files, removed background traffic, and discarded flows with fewer than 10 packets. This curation resulted in 64k samples, which were then used to create five 80/10/10 train/validation/test splits. For SOTA comparison, we rely on results reported by Wang et al.~\cite{wang2024augmentation}. Even though the authors do not explicitly mention using \textit{tcbench} data, their dataset curation, preprocessing steps, and splits are identical to those of \textit{tcbench}. To be certain, we contacted the authors of both~\cite{wang2024augmentation} and \textit{tcbench}, who confirmed it. Therefore, their experiments and ours are based on identical data, making the results suitable for direct comparison.

\subsubsection*{MIRAGE22}
Guarino et al.~\cite{dataset_MIRAGE22} introduced this mobile traffic dataset in 2022, focusing on video meeting apps like Zoom, Webex, and Teams. It includes traffic from nine Android applications. As the name suggests, MIRAGE22 comes from the same research group as MIRAGE19, and therefore the \textit{tcbench} curation process is identical. For SOTA comparison, we rely on the best results reported by Wang et al.~\cite{wang2024augmentation}.

\subsubsection*{UTMOBILENET21}
A mobile traffic dataset containing 17 Android applications, with user interactions emulated through the Android API. Heng et al.~\cite{dataset_UTMOBILENET21} published this dataset in 2021, providing packet information in CSV format. The authors of \textit{tcbench} cleaned the data, assembled flows, filtered flows with less than 10 packets (9.5k samples remained), and prepared five 80/10/10 train/validation/test splits. For SOTA comparison, we rely on results reported by Finamore et al.~\cite{finamore2023replication}, which is the paper that introduced the \textit{tcbench} framework. We can thus be sure that experiments were performed on identical data and that the results are suitable for direct comparison.

\subsubsection*{UCDAVIS19}
A small dataset published in 2019~\cite{dataset_UCDAVIS19} containing QUIC traffic of five Google services: Google Drive, Google Docs, Google Search, Google Music, and  YouTube. It includes a \textit{pretraining} partition with 6.5k samples and two test sets: \textit{human} (83 samples) and \textit{script} (150 samples). Although the small test sets limit the dataset's representativeness, we chose to include it in our evaluation as it is readily available in \textit{tcbench}. We opted not to use the prepared splits, which consist of class-balanced subsets of the \textit{pretraining} partition with only 100 samples per class. Instead, we create ten stratified 80/20 train/validation splits. For SOTA comparison, we rely on results reported by Finamore et al.~\cite{finamore2023replication}.
 
\subsubsection*{CESNET-TLS22}
A large TLS traffic dataset collected from the backbone lines of CESNET, the Czech national research and education network. The dataset spans two weeks and contains 141 million flows categorized into 191 web service classes. Luxemburk et al.~\cite{Luxemburk2023TLS}, the dataset authors, used the first week for training and the second week for testing---a time-based train-test split that we adopt in our experiments. Ten splits are created, each consisting of 1M training flows and 100k validation flows sampled from the first week, and 1M test flows from the second week. For SOTA comparison, we rely on results reported by Fauvel et al.~\cite{fauvel2023lightweight}.

\subsubsection*{AppClassNet}
A large traffic dataset published in 2022 by Wang et al.~\cite{dataset_AppClassNet} that contains 500 applications. It includes two official splits: one for the top 200 applications and another for the remaining 300. We used the top-200 split as follows: the original training and validation sets were merged and re-split into ten stratified train/validation partitions (while keeping the test set unchanged). To ensure reasonable experiment runtimes, each split was then downsampled to 1M training flows, 100k validation flows, and 1M test flows. Before the release of AppClassNet, its packet sequences were transformed to protect business-sensitive information: feature amplitudes were modified, and sequences were partially shuffled. Wang et al.~\cite{dataset_AppClassNet} evaluated several models on both the original and public (transformed) versions of the dataset; for SOTA comparison, we use the results reported on the public version.

\subsection{Transfer learning results}
This section presents transfer learning results obtained on ten downstream TC tasks.~\secref{sec:ranking-transfer-methods} and~\tabref{tab:transfer-methods-results} compare the three transfer methods with each other and with training from scratch, whereas~\secref{sec:surpassing-sota} and~\tabref{tab:cross-eval-results} benchmark the best-performing transfer method and the input-space baseline against the SOTA. Finally,~\secref{sec:input-space-transfer-results} discusses the unexpectedly strong performance of the input-space baseline.

\subsubsection{Ranking of transfer methods}
\label{sec:ranking-transfer-methods}
The results in~\tabref{tab:transfer-methods-results} show a consistent performance ranking among the transfer methods: \textit{linear probing} \textless~\textit{k-NN transfer} \textless~\textit{fine-tuned model}. This order holds across all downstream tasks except for the two UCDAVIS19 test sets. When we compare the fine-tuned model---the best transfer approach---with training from scratch, it performs better on all tasks except for AppClassNet and UCDAVIS19 \textit{script} test, with an average improvement of 2.1\%. Due to its feature transformation process, AppClassNet is a hard downstream task where training from scratch is expected to perform better. In the case of UCDAVIS19, which is the smallest dataset evaluated, the application of the LDIFS regularization technique led to a fine-tuned model that avoided overfitting compared to training from scratch. As a result, it performed much better on the \textit{human} test (+11.57\%, which comes from a different distribution than the training set) but underperformed on the \textit{script} test ($-$0.6\%, which shares the same distribution as the training set). 

While fine-tuning is the overall best approach, we argue that the k-NN transfer method remains valuable in scenarios with limited downstream training data or when fine-tuning is not viable for other reasons. It outperforms training from scratch on four tasks and even surpasses SOTA on all tasks except CESNET-TLS22 and AppClassNet. The strong performance of k-NN transfer demonstrates that even the original, non-fine-tuned embeddings generalize well across diverse TC tasks. 

\begin{table*}[!ht]
    \centering
    \setlength\extrarowheight{1.2pt}
    \caption{Comparison of the best transfer learning method (fine-tuned model) with sota and the input-space baseline. Deltas are computed relative to sota performance. }
    \label{tab:cross-eval-results}
\begin{threeparttable}
\begin{NiceTabular}{|l|c|c|l|lr|lr|c|c|}
\hline
\multicolumn{3}{|c|}{\textbf{Dataset}} &
\multicolumn{5}{c|}{\textbf{Downstream Task Performance [\%]}}  &
\multicolumn{2}{c|}{\textbf{State of the Art }}  
  \\ \hline
%-----------------------------------------------------------
\textbf{Name} & 
\textbf{Subtask} &
\textbf{Classes} &
\multicolumn{1}{c|}{\textbf{SOTA}} &
\multicolumn{2}{c|}{\textbf{Input Space}} &
\multicolumn{2}{c|}{\textbf{Best Transfer}} &
\textbf{Reference} &
\textbf{Year} \\ \hline
%-----------------------------------------------------------
ISCXVPN2016&
Application& 
15&
63.92 & 
\cellcolor[HTML]{FFFFFF} 70.91 & 
\cellcolor[HTML]{57BB8A} \textit{($\Delta$\,\,~~6.99)} & 
\cellcolor[HTML]{FFFFFF} 77.44 & 
\cellcolor[HTML]{57BB8A} \textit{($\Delta$\,~13.52)} & 
Rezaei et al.~\cite{Rezaei2020Multitask} & 
2020 \\ \hline
%-----------------------------------------------------------
ISCXVPN2016&
Traffic type& 
6&
65.56 & 
\cellcolor[HTML]{FFFFFF} 73.01 & 
\cellcolor[HTML]{57BB8A} \textit{($\Delta$\,\,~~7.45)} &
\cellcolor[HTML]{FFFFFF} 80.19 & 
\cellcolor[HTML]{57BB8A} \textit{($\Delta$\,~14.63)} & 
Rezaei et al.~\cite{Rezaei2020Multitask} & 
2020 \\ \hline
%-----------------------------------------------------------
ISCXVPN2016&
Encapsulation& 
2&
85.45 & 
\cellcolor[HTML]{FFFFFF} 90.38 & 
\cellcolor[HTML]{57BB8A} \textit{($\Delta$\,\,~~4.93)} &
\cellcolor[HTML]{FFFFFF} 94.37 & 
\cellcolor[HTML]{57BB8A} \textit{($\Delta$\,\,~~8.92)} & 
Rezaei et al.~\cite{Rezaei2020Multitask} & 
2020 \\ \hline
%-----------------------------------------------------------
MIRAGE19 &
&
20&
80.06\tnote{$\dag$} &
\cellcolor[HTML]{FFFFFF} 79.93 &
\cellcolor[HTML]{FFE5E5} \textit{($\Delta$\:$-$0.13)}   &
\cellcolor[HTML]{FFFFFF} 85.3 &
\cellcolor[HTML]{57BB8A} \textit{($\Delta$\,\,~~5.24)} &
Wang et al.~\cite{wang2024augmentation} &
2024 \\ \hline
%-----------------------------------------------------------
MIRAGE22 &
& 
9&
97.18\tnote{$\dag$} &
\cellcolor[HTML]{FFFFFF} 95.63 &
\cellcolor[HTML]{FFE5E5} \textit{($\Delta$\:$-$1.55)}   &
\cellcolor[HTML]{FFFFFF} 98.67 &
\cellcolor[HTML]{DDF2E8} \textit{($\Delta$\,\,~~1.49)} &
Wang et al.~\cite{wang2024augmentation}        &
2024  \\ \hline
%-----------------------------------------------------------
UTMOBILENET21 &
&
17&
81.91\tnote{$\dag$} & 
\cellcolor[HTML]{FFFFFF} 83.86 & 
\cellcolor[HTML]{DDF2E8} \textit{($\Delta$\,\,~~1.95)} & 
\cellcolor[HTML]{FFFFFF} 88.73 & 
\cellcolor[HTML]{57BB8A} \textit{($\Delta$\,\,~~6.82)} &  
Finamore et al.~\cite{finamore2023replication} & 
2023 \\ \hline
%-----------------------------------------------------------
UCDAVIS19&
\textit{Script} test& 
5&
98.63 & 
\cellcolor[HTML]{FFFFFF} 97.87 & 
\cellcolor[HTML]{FFE5E5} \textit{($\Delta$\:$-$0.76)} & 
\cellcolor[HTML]{FFFFFF} 98.13 & 
\cellcolor[HTML]{FFE5E5} \textit{($\Delta$\:\:\:$-$0.5)} & 
Finamore et al.~\cite{finamore2023replication} & 
2023 \\ \hline
%-----------------------------------------------------------
UCDAVIS19 &
\textit{Human} test&
5&
80.45 &
\cellcolor[HTML]{FFFFFF} 72.77 &
\cellcolor[HTML]{FFB8B8} \textit{($\Delta$\:$-$7.68)}  &
\cellcolor[HTML]{FFFFFF} 89.64 &
\cellcolor[HTML]{57BB8A} \textit{($\Delta$\,\,~~9.19)} &
Finamore et al.~\cite{finamore2023replication} &
2023 \\ \hline
%-----------------------------------------------------------
CESNET-TLS22  &
& 
191&
97.2        &
\cellcolor[HTML]{FFFFFF} 90.95 &
\cellcolor[HTML]{FFB8B8} \textit{($\Delta$\:$-$6.25)}   &
\cellcolor[HTML]{FFFFFF} 98.25 &
\cellcolor[HTML]{DDF2E8} \textit{($\Delta$\,\,~~1.05)} &
Fauvel et al.~\cite{fauvel2023lightweight}     &
2023 \\ \hline 
%-----------------------------------------------------------
AppClassNet  &
& 
200&
88.3        &
\cellcolor[HTML]{FFFFFF} 76.2 &
\cellcolor[HTML]{FFB8B8} \textit{($\Delta$\:$-$12.1)}   &
\cellcolor[HTML]{FFFFFF} 92.11 &
\cellcolor[HTML]{DDF2E8} \textit{($\Delta$\,\,~~3.81)} &
Wang et al.~\cite{dataset_AppClassNet}  &
2022 \\ \hline 
\end{NiceTabular}

\begin{tablenotes}
\leftskip=0.8cm

\item Results are averages across ten splits. For \textit{tcbench} datasets with five prepared splits, we used each split twice.

\item[$\dag$] SOTA is reported in weighted F1-score; therefore, our results for these datasets are also reported using this metric. Nevertheless, we found the differences between the weighted F1-scores and classification accuracies are negligible: \textit{Best Transfer} accuracies MIRAGE19 85.31\%, MIRAGE22 98.67\%, UTMOBILENET21 88.9\%. All results for other datasets are classification accuracies.

\end{tablenotes}
\end{threeparttable}
\end{table*}

\subsubsection{Surpassing SOTA}
\label{sec:surpassing-sota}
We established that fine-tuning the entire model is the most effective transfer method. \tabref{tab:cross-eval-results} compares this approach to SOTA performance, showing that it surpasses it on nine of the ten downstream TC tasks, with an average improvement of 6.4\%. In particular, strong gains exceeding 5\% were achieved on the ISCXVPN2016, MIRAGE19, UTMOBILENET21, and UCDAVIS19 datasets.

The performance on ISCXVPN2016 warrants a closer examination. For SOTA comparison, we used a 1D CNN model of Rezaei et al.~\cite{Rezaei2020Multitask}, which does not utilize payload as input. Nevertheless, our results are close to---or even surpass---the strongest SOTA method that \textit{does} utilize payload data, namely \texttt{DISTILLER-Embeddings} from Nascita et al.~\cite{Nascita2023Embeddings} (results provided in Appendix\ref{ap:sota-comparison}). The performance deltas of our fine-tuned model relative to \texttt{DISTILLER-Embeddings} are as follows: applications~$\Delta$$-$2.48\%, traffic types~$\Delta$$-$1.52\%, encapsulation~$\Delta$+1.36\%. This comparison provides a better estimate of the actual value of including payload as model input on ISCXVPN2016 and shows that packet-sequence-processing and payload-processing models are much closer in performance than previously suggested.

\subsubsection{Surprising performance of the input-space baseline}
\label{sec:input-space-transfer-results}
Among the transfer approach, the input-space baseline, and SOTA methods, we initially expected the input-space baseline to perform the worst due to its absolute simplicity. Surprisingly, however, it outperforms SOTA on four tasks (ISCXVPN2016 tasks and UTMOBILENET21), delivers worse yet comparable performance on three (MIRAGE19, MIRAGE22, and UCDAVIS \textit{script}), and falls significantly behind only on the remaining three (CESNET-TLS22, AppClassNet, and UCDAVIS \textit{human}). To the best of our knowledge, this intriguing finding---that a simple k-NN classifier using features from the first 10 packets performs comparably to SOTA on several datasets---has not been reported in prior work and remains largely unexplored within the TC research domain. We emphasize that no dataset-specific modifications were made, such as adjusting the number of packets or the IPT scaling factor. If the baseline were tuned for each dataset, its performance could likely be improved even further.

We believe that the underlying cause is the high data redundancy in TC datasets. During dataset collection, it is highly probable---or almost certain for script-generated datasets---that multiple instances of the same network communication are captured, such as repeated API requests sent to the same server with identical TCP and TLS configurations. Consequently, TC datasets often contain numerous near-duplicate samples with the same label. When such a dataset is randomly split into training and test sets, these duplicates can end up in both sets. In this scenario, it is not surprising that a classifier relying on the closest training sample achieves high performance.

\section{Limitations}
\label{sec:limitations}

\paragraph*{Nearest neighbors search}
We want to address potential performance concerns related to our use of nearest neighbors search for the domain recognition task. In~\secref{sec:ablation-ranking-speed}, we measured the ranking speed across different embedding sizes. For instance, with an embedding size of 128, the ranking of a database containing one million samples achieves a speed of 10k samples per second. This speed is made possible thanks to \textit{faiss}, which provides efficient methods for finding nearest neighbors and can even run on GPUs for faster processing. In this work, we opted to use the \texttt{IndexFlatIP} index that provides exact ranking results. However, \textit{faiss} also offers other indexes that provide a trade-off between ranking speed and the precision of nearest neighbors search. Thus, if higher ranking speeds were needed, a natural solution would be to use an index with faster ranking, such as \texttt{IndexIVFFlat}\footnotemark, at the cost of losing the guarantee of exact and exhaustive results. 

\paragraph*{Database construction}
A further opportunity for improvement lies in the database construction process. In this work, we used semi-balanced sampling to select one million samples for the database. We believe that a more informative strategy for choosing database samples---for instance, one guided by clustering---could reduce the database size while preserving domain recognition performance by prioritizing highly discriminative samples.

\paragraph*{More research on the input-space baseline needed}
Our motivation for designing a simple baseline and evaluating it under the same conditions as the $\Phi$ embedding function was to establish a reference point. The results, however, turned out to be far more intriguing than anticipated. In the cross-dataset evaluation, the baseline achieved performance quite close to SOTA, even surpassing it for the ISCXVPN2016 and UTMOBILENET21 datasets. Our hypothesis is that the data redundancy inherent in TC datasets, when combined with random splitting into training and test subsets, makes classification trivial for nearest neighbors search. As this observation may influence best practices for constructing and splitting TC datasets, further research is required to rigorously verify it.

\section{Conclusion}
\label{sec:conclusion}
The main objective of this work was to design a universal embedding function suitable for a wide range of TC tasks. We first developed the $\Phi$ embedding function for the exact domain recognition task on the CESNET-QUIC22 dataset and then evaluated how well it generalizes to other TC datasets. To summarize the core components of the proposed embedding function: \textit{(a)} the CNN-based feature extractor that embeds packet features before processing with ResNet-like blocks; \textit{(b)}~the ArcFace loss, which enhances class separation by pulling samples toward class centers while enforcing angular margins; and \textit{(c)} a nearest neighbors classifier using cosine distance in the embedding space.

\footnotetext{\url{https://github.com/facebookresearch/faiss/wiki/Faiss-indexes}.}

The domain recognition task is a significant challenge, particularly when monitoring networks with high volumes of TLS and QUIC traffic protected by the ECH extension, which encrypts entire ClientHello messages and hides SNI domains from network operators. We tackled this task in a setup where domain names were disjoint across training, validation, and test sets, which forced the embedding function to learn traffic patterns that generalize to unseen domains. After an extensive hyperparameter search and tuning of our training pipeline and model architecture, we achieved a classification accuracy of 94.83\% and a recall of 79.35\%, which we consider a strong outcome considering the difficulty of the disjoint-class setup. The proposed architecture maintained a high throughput of 33k embedding inferences per second on a single Nvidia Tesla T4. We also conducted six ablations, each focusing on a specific component to assess its contribution to the overall solution. Notably, the combination of training and database semi-balancing samplers, along with the PLE initialization method for packet size and IPT \textit{Embedding} layers, proved crucial for achieving high recall.

We then transferred the $\Phi$ embedding function to seven TC datasets: ISCXVPN2016, MIRAGE19, MIRAGE22, UTMOBILENET21, UCDAVIS19, CESNET-TLS22, and AppClassNet. The transfer learning approach proved to be highly successful, beating SOTA performance on nine of ten downstream TC tasks. Strong gains exceeding 5\% were achieved on the ISCXVPN2016, MIRAGE19, UTMOBILENET21, and UCDAVIS19 datasets. Overall, the embedding function demonstrated strong generalization across all tested tasks. To our knowledge, no similar transfer learning achievements have been reported in the TC domain. We conclude that the domain recognition task, on which we developed and trained the embedding function, is well suited for the pretraining of TC models due to its complexity, large number of classes, and straightforward labeling process. An additional experiment presented in Appendix~\ref{ap:pretraining-matrix} evaluates individual TC tasks for pretraining, and the results indicate that domain recognition clearly outperforms the other tasks.

\subsection{Future directions}
We conclude this paper by outlining future directions and discussing the advantages of leveraging a network flow’s neighborhood in the embedding space. Producing a ranked list of the N closest samples---i.e., the most similar flows previously observed---offers flexibility in how the results are processed into final predictions. For example, server IP addresses or AS numbers could be used for additional post-filtering of the neighborhood. We see this as a reasonable direction for combining IP-related information with traffic shape characteristics. The ranking output also includes distance values that can support out-of-distribution detection: if all nearest samples exceed a predefined threshold, the prediction could be rejected or adjusted, for example by selecting the most common second-level domain in the neighborhood instead of the full domain prediction. Moreover, predictions based on the most similar samples are inherently more interpretable, as they are accompanied by concrete examples, even though the embeddings themselves are produced by a black-box neural network. Finally, we believe that flow embeddings hold promise for other network monitoring tasks such as device profiling and identification.

\bibliographystyle{IEEEtran}
\bibliography{references.bib}

@IEEEtranBSTCTL{IEEEexample:BSTcontrol,
  CTLdash_repeated_names = "no",
}

@article{Yang2021,
    url = {https://doi.org/10.1109/tnsm.2021.3122940},
    year = {2021},
    month = dec,
    publisher = {Institute of Electrical and Electronics Engineers ({IEEE})},
    volume = {18},
    number = {4},
    pages = {4103--4118},
    author = {Lixuan Yang and Alessandro Finamore and Feng Jun and Dario Rossi},
    title = {Deep Learning and Zero-Day Traffic Classification: Lessons Learned From a Commercial-Grade Dataset},
    journal = {{IEEE} Transactions on Network and Service Management}
}

@article{Luxemburk2023TLS,
    doi = {10.1016/j.comnet.2022.109467},
    url = {https://doi.org/10.1016/j.comnet.2022.109467},
    year = {2023},
    month = jan,
    publisher = {Elsevier {BV}},
    volume = {220},
    pages = {109467},
    author = {Jan Luxemburk and Tom{\'{a}}{\v{s}} {\v{C}}ejka},
    title = {Fine-grained {TLS} services classification with reject option},
    journal = {Computer Networks}
}

@inproceedings{Luxemburk2023QUIC,
    author={Luxemburk, Jan and Hynek, Karel and Čejka, Tomáš},
    booktitle={2023 7th Network Traffic Measurement and Analysis Conference (TMA)}, 
    title={Encrypted traffic classification: the {QUIC} case}, 
    year={2023},
    volume={},
    number={},
    pages={1-10},
    doi={10.23919/TMA58422.2023.10199052}
}

@article{Luxemburk2023QUICDataset,
    doi = {10.1016/j.dib.2023.108888},
    url = {https://doi.org/10.1016/j.dib.2023.108888},
    year = {2023},
    month = feb,
    publisher = {Elsevier {BV}},
    volume = {46},
    pages = {108888},
    author = {Jan Luxemburk and Karel Hynek and Tom{\'{a}}{\v{s}} {\v{C}}ejka and Andrej Luka{\v{c}}ovi{\v{c}} and Pavel {\v{S}}i{\v{s}}ka},
    title = {{CESNET}-{QUIC}22: A large one-month {QUIC} network traffic dataset from backbone lines},
    journal = {Data in Brief}
}

@inproceedings{Deng2019ArcFace,
    author={Deng, Jiankang and Guo, Jia and Xue, Niannan and Zafeiriou, Stefanos},
    booktitle={2019 IEEE/CVF Conference on Computer Vision and Pattern Recognition (CVPR)}, 
    title={{ArcFace}: Additive Angular Margin Loss for Deep Face Recognition}, 
    year={2019},
    volume={},
    number={},
    pages={4685-4694},
    doi={10.1109/CVPR.2019.00482}
}

@inproceedings{Deng2020Subcenter,
    author = {Deng, Jiankang and Guo, Jia and Liu, Tongliang and Gong, Mingming and Zafeiriou, Stefanos},
    title = {Sub-center {ArcFace}: Boosting Face Recognition by Large-Scale Noisy Web Faces},
    year = {2020},
    isbn = {978-3-030-58620-1},
    publisher = {Springer-Verlag},
    address = {Berlin, Heidelberg},
    url = {https://doi.org/10.1007/978-3-030-58621-8_43},
    doi = {10.1007/978-3-030-58621-8_43},
    booktitle = {Computer Vision – ECCV 2020: 16th European Conference, Glasgow, UK, August 23–28, 2020, Proceedings, Part XI},
    pages = {741–757},
    numpages = {17},
    location = {Glasgow, United Kingdom}
}

@misc{Ha2020GoogleLandmark,
    title={Google Landmark Recognition 2020 Competition Third Place Solution}, 
    author={Qishen Ha and Bo Liu and Fuxu Liu and Peiyuan Liao},
    year={2020},
    eprint={2010.05350},
    archivePrefix={arXiv},
    primaryClass={cs.CV},
    url={https://arxiv.org/abs/2010.05350}, 
}

@inproceedings{He2016DeepResidual,
    author={He, Kaiming and Zhang, Xiangyu and Ren, Shaoqing and Sun, Jian},
    booktitle={2016 IEEE Conference on Computer Vision and Pattern Recognition (CVPR)}, 
    title={Deep Residual Learning for Image Recognition}, 
    year={2016},
    volume={},
    number={},
    pages={770-778},
    doi={10.1109/CVPR.2016.90}
}

@misc{Radenovic2018GeM,
    title={Fine-tuning {CNN} Image Retrieval with No Human Annotation}, 
    author={Filip Radenović and Giorgos Tolias and Ondřej Chum},
    year={2018},
    eprint={1711.02512},
    archivePrefix={arXiv},
    primaryClass={cs.CV},
    url={https://arxiv.org/abs/1711.02512}, 
}

@misc{dataset_UCDAVIS19,
    title={How to Achieve High Classification Accuracy with Just a Few Labels: A Semi-supervised Approach Using Sampled Packets}, 
    author={Shahbaz Rezaei and Xin Liu},
    year={2020},
    eprint={1812.09761},
    archivePrefix={arXiv},
    primaryClass={cs.NI},
    url={https://arxiv.org/abs/1812.09761}, 
}

@article{dataset_UTMOBILENET21,
    author={Heng, Yuqiang and Chandrasekhar, Vikram and Andrews, Jeffrey G.},
    journal={IEEE Networking Letters}, 
    title={{UTMobileNetTraffic2021}: A Labeled Public Network Traffic Dataset}, 
    year={2021},
    volume={3},
    number={3},
    pages={156-160},
    doi={10.1109/LNET.2021.3098455}
}

@inproceedings{dataset_MIRAGE19,
    author={Aceto, Giuseppe and Ciuonzo, Domenico and Montieri, Antonio and Persico, Valerio and Pescapé, Antonio},
    booktitle={2019 4th International Conference on Computing, Communications and Security (ICCCS)}, 
    title={{MIRAGE}: Mobile-app Traffic Capture and Ground-truth Creation}, 
    year={2019},
    volume={},
    number={},
    pages={1-8},
    doi={10.1109/CCCS.2019.8888137}
}

@article{dataset_MIRAGE22,
    title = {Contextual counters and multimodal Deep Learning for activity-level traffic classification of mobile communication apps during {COVID-19} pandemic},
    journal = {Computer Networks},
    volume = {219},
    pages = {109452},
    year = {2022},
    issn = {1389-1286},
    doi = {https://doi.org/10.1016/j.comnet.2022.109452},
    author = {Idio Guarino and Giuseppe Aceto and Domenico Ciuonzo and Antonio Montieri and Valerio Persico and Antonio Pescapè},
}

@inproceedings{dataset_ISCXVPN2016,
    title = {Characterization of {{Encrypted}} and {{VPN Traffic}} Using {{Time-related Features}}:},
    shorttitle = {Characterization of {{Encrypted}} and {{VPN Traffic}} Using {{Time-related Features}}},
    booktitle = {Proceedings of the 2nd {{International Conference}} on {{Information Systems Security}} and {{Privacy}}},
    author = {{Draper-Gil}, Gerard and Lashkari, Arash Habibi and Mamun, Mohammad Saiful Islam and A. Ghorbani, Ali},
    year = {2016},
    pages = {407--414},
    doi = {10.5220/0005740704070414},
    isbn = {978-989-758-167-0},
}

@article{dataset_AppClassNet,
      title     = {{AppClassNet}: A commercial-grade dataset for application identification research},
      author    = {Wang, Chao and Finamore, Alessandro and Yang, Lixuan and Fauvel, Kevin and Rossi, Dario},
      journal   = {ACM SIGCOMM Computer Communication Review},
      volume    = {52},
      number    = {3},
      pages     = {19--27},
      year      = {2022},
      month     = {Jul},
      doi       = {10.1145/3561954.3561958}
}

@misc{dataset_AppClassNet_figshare,
    author = "Dario Rossi",
    title = "{AppClassNet - A commercial-grade dataset for application identification research}",
    year = "2022",
    month = "8",
    url = "https://figshare.com/articles/dataset/AppClassNet_-_A_commercial-grade_dataset_for_application_identification_research/20375580",
    doi = "10.6084/m9.figshare.20375580.v1"
}

@inproceedings{Schroff_2015,
    title={FaceNet: A unified embedding for face recognition and clustering},
    url={http://dx.doi.org/10.1109/CVPR.2015.7298682},
    DOI={10.1109/cvpr.2015.7298682},
    booktitle={2015 IEEE Conference on Computer Vision and Pattern Recognition (CVPR)},
    publisher={IEEE},
    author={Schroff, Florian and Kalenichenko, Dmitry and Philbin, James},
    year={2015},
    month=jun, pages={815–823}
}

@inproceedings{Chopra_2005,
    author={Chopra, S. and Hadsell, R. and LeCun, Y.},
    booktitle={2005 IEEE Computer Society Conference on Computer Vision and Pattern Recognition (CVPR'05)}, 
    title={Learning a similarity metric discriminatively, with application to face verification}, 
    year={2005},
    volume={1},
    number={},
    pages={539-546 vol. 1},
    doi={10.1109/CVPR.2005.202}
}

@inproceedings{Wang_2018,
    author={Wang, Hao and Wang, Yitong and Zhou, Zheng and Ji, Xing and Gong, Dihong and Zhou, Jingchao and Li, Zhifeng and Liu, Wei},
    booktitle={2018 IEEE/CVF Conference on Computer Vision and Pattern Recognition}, 
    title={{CosFace}: Large Margin Cosine Loss for Deep Face Recognition}, 
    year={2018},
    volume={},
    number={},
    pages={5265-5274},
    doi={10.1109/CVPR.2018.00552}
}

@misc{sablayrolles2019spreadingvectorssimilaritysearch,
    title={Spreading vectors for similarity search}, 
    author={Alexandre Sablayrolles and Matthijs Douze and Cordelia Schmid and Hervé Jégou},
    year={2019},
    eprint={1806.03198},
    archivePrefix={arXiv},
    primaryClass={stat.ML},
    url={https://arxiv.org/abs/1806.03198}, 
}

@inproceedings{gorishniy2022embeddings,
    title={On Embeddings for Numerical Features in Tabular Deep Learning},
    author={Yury Gorishniy and Ivan Rubachev and Artem Babenko},
    booktitle={{NeurIPS}},
    year={2022},
}

@article{johnson2019billion,
    title={Billion-scale similarity search with {GPUs}},
    author={Johnson, Jeff and Douze, Matthijs and J{\'e}gou, Herv{\'e}},
    journal={IEEE Transactions on Big Data},
    volume={7},
    number={3},
    pages={535--547},
    year={2019},
    publisher={IEEE}
}

@inproceedings{finamore2023replication,
    author = {Finamore, Alessandro and Wang, Chao and Krolikowski, Jonatan and Navarro, Jose M. and Chen, Fuxing and Rossi, Dario},
    title = {Replication: Contrastive Learning and Data Augmentation in Traffic Classification Using a {FlowPic} Input Representation},
    year = {2023},
    isbn = {9798400703829},
    publisher = {Association for Computing Machinery},
    doi = {10.1145/3618257.3624820},
    pages = {36–51},
    numpages = {16},
    location = {Montreal QC, Canada},
    series = {IMC '23}
}

@inproceedings{wang2024augmentation,
    title = {Data Augmentation for Traffic Classification},
    ISBN = {9783031562495},
    ISSN = {1611-3349},
    url = {http://dx.doi.org/10.1007/978-3-031-56249-5_7},
    doi = {10.1007/978-3-031-56249-5_7},
    booktitle = {Passive and Active Measurement},
    publisher = {Springer Nature Switzerland},
    author = {Wang,  Chao and Finamore,  Alessandro and Michiardi,  Pietro and Gallo,  Massimo and Rossi,  Dario},
    year = {2024},
    pages = {159–186}
}

@inproceedings{fauvel2023lightweight,
    author = {Fauvel, Kevin and Chen, Fuxing and Rossi, Dario},
    title = {A Lightweight, Efficient and Explainable-by-Design Convolutional Neural Network for Internet Traffic Classification},
    year = {2023},
    isbn = {9798400701030},
    publisher = {Association for Computing Machinery},
    address = {New York, NY, USA},
    doi = {10.1145/3580305.3599762},
    pages = {4013–4023},
    numpages = {11},
    location = {Long Beach, CA, USA},
    series = {KDD '23}
}

@inproceedings{Luxemburk2023Datazoo,
    author = {Luxemburk, Jan and Hynek, Karel},
    title = {{DataZoo}: Streamlining Traffic Classification Experiments},
    year = {2023},
    isbn = {9798400704499},
    publisher = {Association for Computing Machinery},
    address = {New York, NY, USA},
    url = {https://doi.org/10.1145/3630050.3630176},
    doi = {10.1145/3630050.3630176},
    booktitle = {Proceedings of the 2023 on Explainable and Safety Bounded, Fidelitous, Machine Learning for Networking},
    pages = {3–7},
    numpages = {5},
    location = {Paris, France},
    series = {SAFE '23}
}

@inproceedings{Guarino2023Many,
  author={Guarino, Idio and Wang, Chao and Finamore, Alessandro and Pescapè, Antonio and Rossi, Dario},
  booktitle={2023 7th Network Traffic Measurement and Analysis Conference (TMA)}, 
  title={Many or Few Samples?: Comparing Transfer, Contrastive and Meta-Learning in Encrypted Traffic Classification}, 
  year={2023},
  volume={},
  number={},
  pages={1-10},
  doi={10.23919/TMA58422.2023.10198965}
}

@article{khosla2020supervised,
    title   = {Supervised Contrastive Learning},
    author  = {Prannay Khosla and Piotr Teterwak and Chen Wang and Aaron Sarna and Yonglong Tian and Phillip Isola and Aaron Maschinot and Ce Liu and Dilip Krishnan},
    journal = {arXiv preprint arXiv:2004.11362},
    year    = {2020},
}

@inproceedings{Horowicz2022flowpic,
    author = {Horowicz, Eyal and Shapira, Tal and Shavitt, Yuval},
    title = {A few shots traffic classification with mini-FlowPic augmentations},
    year = {2022},
    isbn = {9781450392594},
    publisher = {Association for Computing Machinery},
    address = {New York, NY, USA},
    url = {https://doi.org/10.1145/3517745.3561436},
    doi = {10.1145/3517745.3561436},
    booktitle = {Proceedings of the 22nd ACM Internet Measurement Conference},
    pages = {647–654},
    numpages = {8},
    location = {Nice, France},
    series = {IMC '22}
}

@article{Nascita2023Embeddings,
    author={Nascita, Alfredo and Montieri, Antonio and Aceto, Giuseppe and Ciuonzo, Domenico and Persico, Valerio and Pescapé, Antonio},
    journal={IEEE Transactions on Network and Service Management}, 
    title={Improving Performance, Reliability, and Feasibility in Multimodal Multitask Traffic Classification with XAI}, 
    year={2023},
    volume={20},
    number={2},
    pages={1267-1289},
    doi={10.1109/TNSM.2023.3246794}
}

@inproceedings{Luxemburk2024_CesnetModels,
    author={Luxemburk, Jan and Hynek, Karel},
    booktitle={2024 8th Network Traffic Measurement and Analysis Conference (TMA)}, 
    title={Towards Reusable Models in Traffic Classification}, 
    year={2024},
    volume={},
    number={},
    pages={1-4},
    doi={10.23919/TMA62044.2024.10559009}
}

@article{Shapira2021_FlowPic,
    author={Shapira, Tal and Shavitt, Yuval},
    journal={IEEE Transactions on Network and Service Management}, 
    title={FlowPic: A Generic Representation for Encrypted Traffic Classification and Applications Identification}, 
    year={2021},
    volume={18},
    number={2},
    pages={1218-1232},
    doi={10.1109/TNSM.2021.3071441}
}

@inproceedings{Xie2023_Rosetta,
    author = {Renjie Xie and Jiahao Cao and Enhuan Dong and Mingwei Xu and Kun Sun and Qi Li and Licheng Shen and Menghao Zhang},
    title = {Rosetta: Enabling Robust {TLS} Encrypted Traffic Classification in Diverse Network Environments with {TCP-Aware} Traffic Augmentation},
    booktitle = {32nd USENIX Security Symposium (USENIX Security 23)},
    year = {2023},
    isbn = {978-1-939133-37-3},
    address = {Anaheim, CA},
    pages = {625--642},
    publisher = {USENIX Association},
    month = aug
}

@article{Mukhoti2024_LDIFS,
    title={Fine-tuning can cripple your foundation model; preserving features may be the solution},
    author={Jishnu Mukhoti and Yarin Gal and Philip Torr and Puneet K. Dokania},
    journal={Transactions on Machine Learning Research},
    issn={2835-8856},
    year={2024},
    url={https://openreview.net/forum?id=kfhoeZCeW7},
}

@inproceedings{Li2018_L2SP,
    title={Explicit Inductive Bias for Transfer Learning with Convolutional Networks}, 
    author={Xuhong Li and Yves Grandvalet and Franck Davoine},
    year={2018},
    booktitle = {International Conference on Learning Representations (ICLR)},
    eprint={1802.01483},
    archivePrefix={arXiv},
    primaryClass={cs.LG},
    url={https://arxiv.org/abs/1802.01483}, 
}

@misc{Dong2022_LRdecay,
    title={CLIP Itself is a Strong Fine-tuner: Achieving 85.7\% and 88.0\% Top-1 Accuracy with ViT-B and ViT-L on ImageNet}, 
    author={Xiaoyi Dong and Jianmin Bao and Ting Zhang and Dongdong Chen and Shuyang Gu and Weiming Zhang and Lu Yuan and Dong Chen and Fang Wen and Nenghai Yu},
    year={2022},
    eprint={2212.06138},
    archivePrefix={arXiv},
    primaryClass={cs.CV},
    url={https://arxiv.org/abs/2212.06138}, 
}

@inproceedings{Rezaei2020Multitask,
    author={Rezaei, Shahbaz and Liu, Xin},
    booktitle={2020 29th International Conference on Computer Communications and Networks (ICCCN)}, 
    title={Multitask Learning for Network Traffic Classification}, 
    year={2020},
    volume={},
    number={},
    pages={1-9},
    doi={10.1109/ICCCN49398.2020.9209652}
}

@article{Nascita2021FirstEmbeddings,
    author={Nascita, Alfredo and Montieri, Antonio and Aceto, Giuseppe and Ciuonzo, Domenico and Persico, Valerio and Pescapé, Antonio},
    journal={IEEE Transactions on Network and Service Management}, 
    title={XAI Meets Mobile Traffic Classification: Understanding and Improving Multimodal Deep Learning Architectures}, 
    year={2021},
    volume={18},
    number={4},
    pages={4225-4246},
    doi={10.1109/TNSM.2021.3098157}
}

@inproceedings{lin2022Etbert,
    author    = {Xinjie Lin and
               Gang Xiong and
               Gaopeng Gou and
               Zhen Li and
               Junzheng Shi and
               Jing Yu},
    title     = {{ET-BERT:} {A} Contextualized Datagram Representation with Pre-training
               Transformers for Encrypted Traffic Classification},
    booktitle = {{WWW} '22: The {ACM} Web Conference 2022, Virtual Event, Lyon, France,
               April 25 - 29, 2022},
    pages     = {633--642},
    publisher = {{ACM}},
    year      = {2022}
}

@inproceedings{Zhao2023Yatc,
    author = {Zhao, Ruijie and Zhan, Mingwei and Deng, Xianwen and Wang, Yanhao and Wang, Yijun and Gui, Guan and Xue, Zhi},
    title = {Yet another traffic classifier: a masked autoencoder based traffic transformer with multi-level flow representation},
    year = {2023},
    isbn = {978-1-57735-880-0},
    publisher = {AAAI Press},
    url = {https://doi.org/10.1609/aaai.v37i4.25674},
    doi = {10.1609/aaai.v37i4.25674},
    booktitle = {Proceedings of the Thirty-Seventh AAAI Conference on Artificial Intelligence and Thirty-Fifth Conference on Innovative Applications of Artificial Intelligence and Thirteenth Symposium on Educational Advances in Artificial Intelligence},
    articleno = {605},
    numpages = {8},
    series = {AAAI'23/IAAI'23/EAAI'23}
}

@article{Bovenzi2024Attack,
    title = {Classifying attack traffic in IoT environments via few-shot learning},
    journal = {Journal of Information Security and Applications},
    volume = {83},
    pages = {103762},
    year = {2024},
    issn = {2214-2126},
    doi = {https://doi.org/10.1016/j.jisa.2024.103762},
    url = {https://www.sciencedirect.com/science/article/pii/S2214212624000656},
    author = {Giampaolo Bovenzi and Davide {Di Monda} and Antonio Montieri and Valerio Persico and Antonio Pescapè},
}

@article{Monda2024Few,
    author={Di Monda, Davide and Montieri, Antonio and Persico, Valerio and Voria, Pasquale and De Ieso, Matteo and Pescapè, Antonio},
    journal={IEEE Open Journal of the Communications Society}, 
    title={Few-Shot Class-Incremental Learning for Network Intrusion Detection Systems}, 
    year={2024},
    volume={5},
    number={},
    pages={6736-6757},
    doi={10.1109/OJCOMS.2024.3481895}
}

@inproceedings{Monda2024Botnet,
    author={Monda, Davide Di and Bovenzi, Giampaolo and Montieri, Antonio and Persico, Valerio and Pescapè, Antonio},
    booktitle={2023 IEEE International Conference on Big Data (BigData)}, 
    title={IoT Botnet-Traffic Classification Using Few-Shot Learning}, 
    year={2023},
    volume={},
    number={},
    pages={3284-3293},
    doi={10.1109/BigData59044.2023.10386602}
}

@misc{Tian2020RFS,
    title={Rethinking Few-Shot Image Classification: a Good Embedding Is All You Need?}, 
    author={Yonglong Tian and Yue Wang and Dilip Krishnan and Joshua B. Tenenbaum and Phillip Isola},
    year={2020},
    eprint={2003.11539},
    archivePrefix={arXiv},
    primaryClass={cs.CV},
    url={https://arxiv.org/abs/2003.11539}, 
}

@article{Tong2025Adaptation,
    author={Tong, Van and Dao, Cuong and Tran, Hai-Anh and Tran, Duc and Thi Thanh Binh, Huynh and Hoang-Nam, Thang and Tran, Truong X.},
    journal={IEEE Transactions on Network and Service Management}, 
    title={Encrypted Traffic Classification Through Deep Domain Adaptation Network With Smooth Characteristic Function}, 
    year={2025},
    volume={22},
    number={1},
    pages={331-343},
    doi={10.1109/TNSM.2025.3534791}
}

@ARTICLE{Gioacchini2024Cross,
    author={Gioacchini, Luca and Mellia, Marco and Vassio, Luca and Drago, Idilio and Milan, Giulia and Houidi, Zied Ben and Rossi, Dario},
    journal={IEEE Transactions on Network and Service Management}, 
    title={Cross-Network Embeddings Transfer for Traffic Analysis}, 
    year={2024},
    volume={21},
    number={3},
    pages={2686-2699},
    doi={10.1109/TNSM.2023.3329442}
}

@inproceedings{Gioacchini2021Darknet,
    author = {Gioacchini, Luca and Vassio, Luca and Mellia, Marco and Drago, Idilio and Houidi, Zied Ben and Rossi, Dario},
    title = {DarkVec: automatic analysis of darknet traffic with word embeddings},
    year = {2021},
    isbn = {9781450390989},
    publisher = {Association for Computing Machinery},
    address = {New York, NY, USA},
    url = {https://doi.org/10.1145/3485983.3494863},
    doi = {10.1145/3485983.3494863},
    booktitle = {Proceedings of the 17th International Conference on Emerging Networking EXperiments and Technologies},
    pages = {76–89},
    numpages = {14},
    location = {Virtual Event, Germany},
    series = {CoNEXT '21}
}

\appendices

\section{State-of-the-art comparison}
\label{ap:sota-comparison}

For each dataset, we reference the prior work used for the SOTA comparison. We specify the evaluation metric and indicate the exact tables from which information on the best-performing classifiers was obtained. In some cases, measurements from multiple tables must be combined to provide the best achieved result reported in the cited work.

{\small
\begin{description}[leftmargin=14pt]
    \item[ISCXVPN2016, accuracy] -- Table 3 of Nascita et al.~\cite{Nascita2023Embeddings}. We primarily compare to a model that does not utilize payload as input, \textit{“1D-CNN (PSQ)”}, which achieves 85.45\% on the Encapsulation task, 65.56\% on the Traffic Type task, and 63.92\% on the Application task. In~\secref{sec:surpassing-sota}, we also compare to the  \texttt{DISTILLER-Embeddings} model, likewise reported in Table 3 of Nascita et al.~\cite{Nascita2023Embeddings}. This payload-processing model achieves 93.01\% on the Encapsulation task, 81.71\% on the Traffic Type task, and 79.92\% on the Application task.    
    \item[MIRAGE19, weighted F1-score] -- Table 7 of Wang et al.~\cite{wang2024augmentation}. Deltas from this table need to be added to the baseline performance of 75.43\%. The best result is \textit{"MaskedStack (p = 0.7)"} with 80.06\% (75.43\% + 4.63\%).
    \item[MIRAGE22, weighted F1-score] -- Table 7 of Wang et al.~\cite{wang2024augmentation}. Deltas from this table need to be added to the baseline performance of 94.92\%. The best result is \textit{"MaskedStack (p = 0.3)"} with 97.18\% (94.92\% + 2.26\%). 
    \item[UTMOBILENET21, weighted F1-score] -- Table 8 of Finamore et al.~\cite{finamore2023replication}. The best result for the \textgreater10pkts version is \textit{"Time shift"} with 81.91\%.
    \item[UCDAVIS19, accuracy] -- Table 7 of Finamore et al.~\cite{finamore2023replication}, which reports results for an enlarged training set. The best result on the \textit{human} test set is \textit{“SimCLR + fine-tuning”} with 80.45\%. For the \textit{script} test set, the best result is \textit{“Packet loss”} with 98.63\%.
    \item[CESNET-TLS22, accuracy] -- Table 5 of Fauvel et al.~\cite{fauvel2023lightweight}. The best result of 97.2\% is achieved with the LEXNet architecture.
    \item[AppClassNet, accuracy] -- Table 3 of Wang et al.~\cite{dataset_AppClassNet}. The best result of 88.3\% on the public version of the dataset is achieved with random forest.
\end{description}
}

\begin{table*}[!ht]
    \centering
    \scriptsize
    \renewcommand{\arraystretch}{1.5} % spacing
    \caption{Downstream task performance using weights trained from scratch on source tasks, replacing the original domain recognition weights. The last row presents linear probing results of the original model from~\tabref{tab:transfer-methods-results}. Results are averages of classification accuracies across five splits.}
    \label{tab:pretraining-downstream-matrix}
    \begin{NiceTabular}{|c|*{10}{c|}}
    \hline
        \multirow{2}{*}{\textbf{Source Task}} &
        \multicolumn{10}{c|}{\textbf{Downstream Task Performance [\%]}} \\ \cline{2-11}
        & \rotatebox{90}{ISCX Application} &
          \rotatebox{90}{ISCX TrafficType} &
          \rotatebox{90}{ISCX Encapsulation} &
          \rotatebox{90}{MIRAGE19} &
          \rotatebox{90}{MIRAGE22} &
          \rotatebox{90}{UTMOBILENET21} &
          \rotatebox{90}{UCDAVIS19 \textit{Script}~~} &
          \rotatebox{90}{UCDAVIS19 \textit{Human}~~} &
          \rotatebox{90}{CESNET-TLS22} &
          \rotatebox{90}{AppClassNet} \\
    \hline
    ISCX Application & \cellcolor[HTML]{E8E8E8} & \cellcolor[HTML]{E8E8E8} & \cellcolor[HTML]{E8E8E8} & 46.14 & 75 & 71.29 & 91.87 & 50.6 & 46.84 & 45.98\\ \hline
    ISCX TrafficType & \cellcolor[HTML]{E8E8E8} & \cellcolor[HTML]{E8E8E8} & \cellcolor[HTML]{E8E8E8} & 46.29 & 79.06 & 72.3 & 90.8 & 54.22 & 47.05 & 39.62\\ \hline
    ISCX Encapsulation & \cellcolor[HTML]{E8E8E8} & \cellcolor[HTML]{E8E8E8} & \cellcolor[HTML]{E8E8E8} & 47.03 & 81.19 & 77.02 & 92.4 & 53.73 & 45.81 & 45.93 \\ \hline
    MIRAGE19 & 58.76 & 60.4 & 79.64 & \cellcolor[HTML]{E8E8E8} & 71.76 & 69.26 & 86.27 & 44.82 & 34.42 & 49.64 \\ \hline
    MIRAGE22 & 60.02 & 62.92 & 83.05 & 48.84 & \cellcolor[HTML]{E8E8E8} & 65.84 & 93.6 & 54.7 & 46.38 & 46.13 \\ \hline
    UTMOBILENET21 & 55.1 & 57.45 & 79.65 & 46.4 & 74.87 & \cellcolor[HTML]{E8E8E8} & 89.07 & 58.8 & 47.22 & 45.54 \\ \hline
    UCDAVIS19 & 40.64 & 49.47 & 74.95 & 39.52 & 72.02 & 70.06 & \cellcolor[HTML]{E8E8E8} & \cellcolor[HTML]{E8E8E8} & 19.72 & 36.34 \\ \hline
    CESNET-TLS22 & 68.81 & 69.67 & 85.29 & 66.17 & 94.22 & 82.58 & 98.27 & 64.1 & \cellcolor[HTML]{E8E8E8} & 60.15 \\ \hline
    AppClassNet & 63.79 & 66.47 & 84.38 & 59.85 & 90.2 & 78.96 & 94 & 61.45 & 62.78 & \cellcolor[HTML]{E8E8E8} \\ \hline
     \cellcolor[HTML]{FF9F40} QUIC Domain Recognition & \cellcolor[HTML]{FF9F40} 72.81 & \cellcolor[HTML]{FF9F40} 73.01 & \cellcolor[HTML]{FF9F40} 88.03 & \cellcolor[HTML]{FF9F40} 72.51 & \cellcolor[HTML]{FF9F40} 95.89 & \cellcolor[HTML]{FF9F40} 86.43 & \cellcolor[HTML]{FF9F40} 98 & \cellcolor[HTML]{FF9F40} 87.47 & \cellcolor[HTML]{FF9F40} 85.65  & \cellcolor[HTML]{FF9F40} 62.15 \\ \hline
    \end{NiceTabular}
\end{table*}

\section{\texorpdfstring{Pretraining $\times$ downstream task matrix}{Pretraining x downstream task matrix}}
\label{ap:pretraining-matrix}

To further assess the strength of domain recognition as a pretraining task, we used individual TC tasks for pretraining instead of domain recognition. Specifically, we used models trained from scratch (those reported in the \textit{From Scratch} column of~\tabref{tab:transfer-methods-results}). We then performed linear probing to evaluate how well these models transfer to other tasks. In other words, for each row in~\tabref{tab:pretraining-downstream-matrix}, we repeated the linear probing experiment described in~\secref{sec:transfer-methods}, but replaced the original model weights obtained via domain recognition with weights trained from scratch on the source task.

We did not perform cross-transfer among the three ISCXVPN2016 tasks, as this would risk data snooping given our experimental setup with separate train/validation/test splits for these tasks. We chose linear probing for this experiment because it provides a straightforward interpretation of the results: it measures how linearly separable the features transferred across TC tasks are.

\subsection{Results}

The results indicate that pretraining on other datasets produces features that do not generalize well. Larger datasets, such as CESNET-TLS22 and AppClassNet, achieve somewhat comparable performance but still lag behind domain recognition pretraining by approximately 5--10\%. Smaller datasets perform poorly as pretraining tasks in this setup.

An open question is whether domain recognition itself constitutes an effective pretraining task, or whether the ArcFace-based training pipeline described in~\secref{sec:experimental-setup} is the primary driver behind the strong transfer performance. Evaluating and tuning the training pipeline for all source tasks is, however, beyond the scope of this paper. Nonetheless, we conclude that the combination of domain recognition---as a challenging task with hundreds of classes---and our optimized training pipeline produces features that are effective across all tested downstream tasks. Based on the preliminary experiments presented here, such effectiveness is difficult to achieve through pretraining on other source tasks.

\end{document}